\documentclass{article}
\include{vision.def}

\usepackage[utf8]{inputenc}
\usepackage[english]{babel}
\usepackage{xcolor}

\usepackage{arxiv}
\usepackage{mathtools}
\usepackage[utf8]{inputenc} 
\usepackage[T1]{fontenc}    
\usepackage{hyperref}       
\usepackage{url}            
\usepackage{booktabs}       
\usepackage{amsfonts}       
\usepackage{nicefrac} 
\usepackage{microtype}
\usepackage{graphicx}
\usepackage{authblk}
\usepackage{caption}
\usepackage{subfigure}
\usepackage{amsmath}
\usepackage{multirow}
\usepackage{array}
\usepackage{verbatim}
\usepackage{datetime}

\usepackage{enumitem}
\setlist[enumerate]{leftmargin=*}
\setlist[itemize]{leftmargin=*}
\setlist[itemize]{leftmargin=6mm}

\newcommand*{\defeq}{\stackrel{\mathsf{def}}{=}}

\title{Markov-Lipschitz Deep Learning}

\author{
Stan Z. Li  \ \ \ \ \ \  Zelin Zang  \ \ \ \ \ \ Lirong Wu\\
AI Lab, School of Engineering, Westlake University\\
Institute of Advanced Technology, Westlake Institute for Advanced Study\\
Hangzhou, Zhejiang, China\\
}

\begin{document}


\maketitle

\begin{abstract}
We propose a novel framework, called Markov-Lipschitz deep learning (MLDL), to tackle geometric deterioration caused by collapse, twisting, or crossing in vector-based neural network transformations for manifold-based representation learning and manifold data generation. A prior constraint, called {\em locally isometric smoothness} (LIS), is imposed across-layers and encoded into a Markov random field (MRF)-Gibbs distribution. This leads to the best possible solutions for local geometry preservation and robustness as measured by locally geometric distortion and locally bi-Lipschitz continuity. Consequently, the layer-wise vector transformations are enhanced into well-behaved,  LIS-constrained metric homeomorphisms. Extensive experiments, comparisons, and ablation study demonstrate significant advantages of MLDL for manifold learning and manifold data generation. MLDL is general enough to enhance any vector transformation-based networks. 
The code is available at  \textcolor{red}{https://github.com/westlake-cairi/Markov-Lipschitz-Deep-Learning}.
\end{abstract}

\section{Introduction}

{\bf Manifold learning} aims to perform nonlinear dimensionality reduction (NLDR) mapping from the input data space to a latent space so that we can use the Euclidean metric to facilitate pattern analysis and {\bf manifold data generation} does the inverse from the learned latent space. Numerous literature exist on manifold learning for NLDR, including classic such as ISOMAP \cite{Tenenbaum-science-00} and LLE (locally linear embedding) \cite{Roweis-science-00} and more recent  developments \cite{donoho2003hessian,hinton2006reducing,zhang2007mlle,CHEN-JASA-2009,Gashler-NIPS-2007,McQueen-NIPS-2016} and popular visualization methods such as t-SNE \cite{maaten2008visualizing}. The problem can be considered from the viewpoints of geometry \cite{Bronstein-GDL-2017}) and topology \cite{Wasserman-2016,Moor19Topological}. The manifold assumption \cite{Belkin-Niyogi-02,Fefferman-manifold-2016} is the basis for NLDR, and preserving local geometric structure is the key to its success.

\begin{center}
\begin{table}[h!]
\centering
\caption{Capability comparison of different methods}
\begin{tabular}{lccccc}
\toprule
   & MLDL  &  AE/TopoAE & MLLE & ISOMAP & t-SNE   \\
\midrule
Manifold learning without decoder               & Yes & No & Yes & Yes & Yes \\
Learned NLDR applicable to test data             & Yes & Yes & No & No & No\\
Able to generate data of learned manifolds   & Yes & No & No & No & No\\
\bottomrule
\end{tabular}
\label{tbl:MLDL}
\end{table}
\end{center}

In this paper, we develop a novel deep learning framework, called Markov-Lipschitz deep learning ({\bf MLDL}), for manifold learning-based NLDR, representation learning and data generation tasks. The motivations are the following: (1) We believe that existing layer-wise vector transformations of neural networks can be enhanced by imposing on them a constraint, which we call {\em locally isometric smoothness} ({\bf LIS}), to become metric {\em homeomorphisms}. The resulting LIS-constrained homeomorphisms are continuous and bijective mappings, and due to the geometry-preserving property of LIS, avoid the mappings from collapse, twisting, or crossing, thus improve generalization, stability, and robustness. (2) Existing manifold learning methods are based on local geometric structures of data samples, thus may be modeled by conditional probabilities of {\em Markov random fields} (MRFs) locally and an MRF-Gibbs distribution globally.  

The paper combines the above two features into the MLDL framework by implementing the LIS constraint through {\em locally bi-Lipschitz continuity} and encoding it into the energy (loss) function of an MRF-Gibbs distribution. The functional features of MLDL are summarized in Table~\ref{tbl:MLDL} in comparison with other popular methods. The proposed MLDL has advantages over other AE-based methods in terms of manifold learning without a decoder and generating new data from the learned manifold. These merits are ascribed to its distinctive property of bijection and invertibility endowed by the LIS constraint. The main contributions, to the best of our knowledge, are summarized below: 


\begin{itemize}
\item[(1)] Proposing the MLDL framework that (i) imposes the prior LIS constraint across network layers, (ii) constrains a neural network as a cascade of homeomorphic transformations, and (iii) encodes the constraint into an MRF-Gibbs prior distribution. This results in MLDL-based neural networks optimized in terms of not only local geometry preservation measured by geometric distortion but also homeomorphic regularity measured by locally bi-Lipschitz constant.

\item[(2)] Proposing two instances of MLDL-based neural networks: Markov-Lipschitz Encoder ({\bf ML-Enc}) for manifold learning and ML AutoEncoder ({\bf ML-AE}). The decoder part ({\bf ML-Dec}) of the ML-AE helps regularize manifold learning of ML-Enc and also acts as a manifold data generator. 

\item[(3)] Proposing an auxiliary term for MLDL training. It assists graduated optimization and prevents MLDL from falling into bad local optima (failure in unfolding the manifold into a plane in latent space).

\item[(4)] Providing extensive experiments, with self- and comparative evaluations and ablation study, which demonstrate significant advantages of MLDL over existing methods in terms of locally geometric distortion and locally bi-Lipschitz continuity.

\end{itemize}

\noindent \textbf{Related work.} Besides the above-mentioned literature, related work also includes the following: Markov random field (MRF) concepts \cite{Geman-Geman-84,Li-MRF-Book-95} are used for establishing a connection between neighborhood-based loss (energy) functions and the global MRF-Gibbs distribution. Lipschitz continuity has been used to improve neural networks in generalization \cite{Bartlett-NIPS-2017,Anil-ICML-2018}, robustness against adversarial attacks \cite{Miyato-ICLR-2018,Weng-ICLR-2019,Cohen-ICML-2019}, and generative modeling \cite{Zhou-Lip-GAN-ICML2019,Qi-LS-GAN-IJCV2020}. Algorithms for estimation of Lipschitz constants are developed in \cite{Virmaux-NIPS-2018,Gulrajani-NIPS-2017,Miyato-ICLR-2018,Zhang-Lip-Estimation-NIPS-2018,Latorre-ICLR-2020}.

\noindent \textbf{Organization.}  In the following, Section~2 introduces the MLDL network structure and related preliminaries, Section~3 presents the key ideas of MLDL and the two MLDL neural networks and Section~4 presents extensive experiments. More details of experiments can be found in Appendix.

\section{MLDL Network Structure}

\begin{figure}[h]
  \centering
  \includegraphics[width=5.5in]{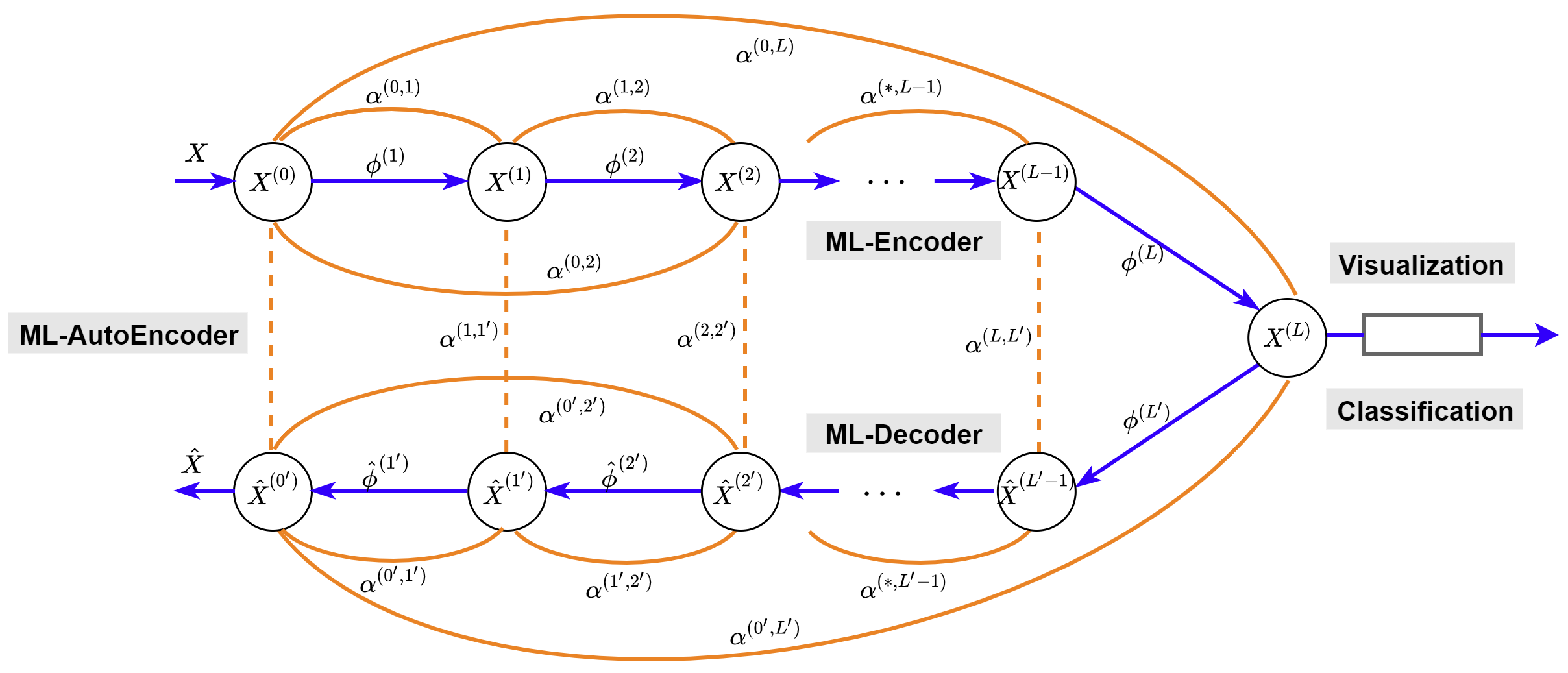}
\caption{Illustration of MLDL framework.
The ML-AutoEncoder transforms the input $X$ to the latent $X^{(L)}$ using the ML-Encoder, and then to reconstructed $\hat X$ using the ML-Decoder (best viewed in color). Whereas a standard neural network consists of a cascade of transformations $\phi^{(l)}$ (blue arrows), an ML-network imposes the LIS constraint between {\bf any} two layers as appropriate (shown in orange arcs and dashed lines)  cross-layer loss functions weighted by $\alpha^{(l,l')}$. This encourages $\phi^{(l)}$ to become well-behaved local homeomorphisms. Latent features $X^{(L)}$ extracted by the learned ML-Encoder can be used for downstream tasks such as visualization and classification as well as manifold data generation using the learned ML-Decoder.} 
  \label{fig:MLDL}
\end{figure}

The structure of MLDL networks is illustrated in Fig.~\ref{fig:MLDL}. The ML-Encoder, composed of a cascade of locally homeomorphic transformations, is aimed at manifold learning and NLDR, and the corresponding ML-Decoder is for manifold reconstruction and data generation. The LIS constraint is imposed across layers and encoded into the energy (loss) function of an MRF-Gibbs distribution. These lead to good properties in local geometry preservation and locally bi-Lipschitz continuity. This section introduces preliminaries and describes the MLDL concepts of local homeomorphisms and the LIS constraint in connection to MLDL.

\subsection{MRFs on neural networks}

Markov random fields (MRFs) can be used for manifold learning-based NLDR and manifold data generation at the data sample level because the modeling therein is done with respect to some neighborhood system. On the other hand, at the neural network level, layer-wise interaction and performance of a neural network can be model by Markov or relational graphs (e.g. \cite{awiszus_markov_2018,You-graph-NN-2020}). This section introduces the basic notion of MRFs for modeling relationships between data at each layer and interactions between layers.

\noindent \textbf{Data samples on manifold.} Let $X=\left\{x_1,\ldots,x_M\right\}$  be a set of $M$ samples on a manifold $\mathcal{M}_X\subset\mathbb{R}^N$. We aim to learn a latent representation of  $\mathcal{M}_X$ from $X\subset\mathcal{M}_X$ based on the local structure of $X$. For this we need to define a metric $d_X$ and a neighborhood system $\mathcal{N}=\left\{\mathcal{N}_{i} \ | \ \forall i\in\mathcal{S}\right\}$. In this paper, we use the Euclidean metric for $d_X$ and define $\mathcal{N}_i$ as the $k$-NN of $x_i$, and when we write a pair $(i,j)\in\mathcal{S}\times\mathcal{S}$, it is restricted by $j\in\mathcal{N}_{i}$ unless specified otherwise. Also, note that the concept of $\mathcal{N}_i$ is used for training an MLDL network only and not needed for testing. When $\mathcal{M}_X$ is Riemannian, its tangent subspace $T_x(\mathcal{M}_X)$ at any $x\in\mathcal{M}_X$ is locally isomorphic to the Euclidean space of $dim(\mathcal{M}_X)$ and this concept is the basis for manifold learning. 

\noindent \textbf{MRF modeling of manifold data samples.} Manifold learning algorithms usually work with a neighborhood system, namely, the direct influence of every sample point $x_i$ on the other points is limited to its neighbors $x_j$ for $j \in \mathcal{N}_i$ only (condition 1). As long as the neighboring relationship is defined, the conditional distribution $p(x_i \ | \ x_j) > 0 $ is positive (condition 2). The collection of random variables $X$ is an MRF when these two conditions are satisfied, which is generally the case for manifold learning algorithms. 

Furthermore, $X$ is an MRF on $\mathcal{S}$ with respect to $\mathcal{N}$ if and only if its joint distribution $P(X) $ is a Gibbs distribution with respect to $\mathcal{N}$ \cite{Besag-74}. A Gibbs distribution (GD) of $X$ is featured by two things: (1) it belongs to the exponential family; (2) its energy function is defined on cliques. A clique $c$ consists of either a single node $\{i\}$ or a pair of neighboring nodes $\{i,j\}$. Let's denote the respective set by $\mathcal{C}_1=\{\{i\}\;|\;i\in\mathcal{S}\}$ and $\mathcal{C}_2\{\{i,j\} \ | \ j \in\mathcal{N}_i,\;i\in\mathcal{S}\}$ and the collection of all given cliques by  $\mathcal{C}=\mathcal{C}_1 \ \cup \ \mathcal{C}_2$ if $\mathcal{C}_1$ is also taken consideration. A GD of $X$ with respect to $\mathcal{N}$ is of the following form
\begin{align*}
p\left(X\right) \propto e^{-U\left(X\right)/T},
\end{align*}
where 
\begin{align}
U(X)=\sum_{c\in \mathcal{C}}V_c(X).
\label{equ:Gibbs-energy-function}
\end{align}
is the energy composed of clique potentials $V_c(X)$, and $T$ is a global parameter called the temperature. The MRF-Gibbs equivalence theorem enables the formulations of loss functions in a principled way based on probability and statistics, rather than heuristically, thus laying a rational foundation for optimization-based manifold learning. The LIS loss and other loss functions to be formulated in this paper are all based on clique potentials-based energy functions as in Equ.~(\ref{equ:Gibbs-energy-function}).

\noindent \textbf{MRF modeling of neural network layers.}
We extend the concepts of the neighborhood system and clique potentials to impose between-layer constraints. Consider an $L$-layered manifold transforming  neural network, such as the ML-Encoder in Fig.~\ref{fig:MLDL}. We have $(X^{(l)},d^{(l)}_X)$ as the metric space for layer $l$. Let  $\widetilde{\mathcal{S}} = \left\{0,\ldots,L\right\}$  and $\widetilde{X}=\left\{X^{(0)},X^{(1)},\ldots,X^{(L)}\right\}$ in which $X^{(0)}=X$ is the input data and fixed, and all the other $X^{(l)}$ (for $l>0$) are part of the transformed data or solution. Between a pair of layers $l$ and $l'$ can exist a link (undirected) which we call {\em super-cliques}. Let $\widetilde{\mathcal{C}}_{ALL}=\left\{\{l,l'\} \ | \ l<l',\ l\in\widetilde{\mathcal{S}}\right\}$ be the set of all super-cliques, and $\widetilde{\mathcal{C}}_2\subseteq\widetilde{\mathcal{C}}_{ALL}$ be the subset of the pairwise super-cliques under consideration. Underlying $\widetilde{\mathcal{C}}_2$ is a neighborhood system $\widetilde{\mathcal{N}}$ which is independent of the input data. 
$\widetilde{X}$ is an MRF on $\widetilde{\mathcal{S}}$ w.r.t. $\widetilde{\mathcal{N}}$ and its global joint probability can be modeled by a clique potential-based Gibbs distribution.

\subsection{Local homeomorphisms}

\textbf{Transformation between graphs.} Manifold learning from $X$ is associated with a graph $\mathcal{G}\left(X,D_X,\mathcal{N}_X\right)$ where $\mathcal{N}_X$ can be defined from the distance matrix $D_X=[d_X(x_i,x_j)]$.  The objective of {\em manifold learning} or NLDR is to find a local homeomorphism
\begin{align*}
\Phi : \mathcal{G}\left(X,D_X,\mathcal{N}_X\right)\rightarrow\ \mathcal{G}\left(Z,D_Z,\mathcal{N}_Z\right),
\end{align*}
transforming from the input space $X\subset\mathcal{M}_X\subset\mathbb{R}^N$ to the latent space $Z\subset\mathcal{M}_Z\subset\mathbb{R}^n$ with $n<N$.
In this work, $\Phi$ is realized by the ML-Enc which is required to best satisfy the LIS constraint. The reverse process, {\em manifold data generation} realized by the ML-Dec, is also a local homeomorphism
\begin{align*}
 \Phi^{-1} : \mathcal{G}\left(Z,D_Z,\mathcal{N}_Z\right)\rightarrow \mathcal{G}\left(X,D_X,\ \mathcal{N}_X\right),
\end{align*}

\noindent \textbf{Cascade of local homeomorphisms.} While $\Phi$ (and its inverse) can be highly nonlinear and complex, we decompose it into  a cascade of $L$ less nonlinear, locally isometric homeomorphisms $\mathrm{\Phi}=\ \phi^{(L)}\circ\cdots\circ\phi^{(2)}\circ {\Phi}^{(1)}$. This can be done using an $L$-layer neural network (e.g. ML-Enc) of the following form
\begin{align*}
\Phi : \mathcal{G}(X^{(0)},D^{(0)}, \mathcal{N}^{(0)}) \stackrel{\phi^{(1)}}{\longrightarrow} 
        \mathcal{G}(X^{(1)},D^{(1)}, \mathcal{N}^{(1)}) \stackrel{\phi^{(2)}}{\longrightarrow}
        \cdots  \stackrel{\phi^{(L)}}{\longrightarrow}
        \mathcal{G}(X^{(L)},D^{(L)}, \mathcal{N}^{(L)}).
\end{align*}
in which $X^{(0)}=X$ is the input data, $\phi^{(l)}$ the nonlinear transformation at layer $l\in\{0,\ldots,L\}$, $X^{(l)}{\subset\mathbb{R}}^{N_l}$ the output of $\phi^{(l)}$, $\mathcal{N}^{(l)}$ the neighborhood system, and $D^{(l)}=[d_l({x_i}^{\left(l\right)}$,${x_j}^{\left(l\right)})]$ the distance matrix given metric $d_l$.  The layer-wise transformation can be written as
\begin{align*}
X^{(l+1)}=\phi^{(l)}\ (X^{(l)}, \ D^{(l)},\ \mathcal{N}^{(l)}\ | \ W^{(l)}).
\end{align*}
in which $W^{(l)}$ is the weight matrix of the neural network $\Phi$ to be learned, the distance matrix $D^{(l)}$ is updated after iterations, and the product $W^{(l)} X^{(l)}$ is followed by a nonlinear activation. 

Such a decomposition is made possible by the property that the tangent space of a Riemannian manifold is locally isomorphic to a simple Euclidean metric space. The layer-wise neural network unfolds $\mathcal{M}_X$ in the input space, stage by stage, into $\mathcal{M}_Z$ in a latent space to best preserve the local geometric structure of $\mathcal{M}_X$. In other words, given $X^{(0)}=X$, $\phi^{(l)}$ transforms $\mathcal{M}^{(l-1)}$ onto $\mathcal{M}^{(l)}{\subset\mathbb{R}}^{N_l}$, finally resulting in the embedding as $Z =X^{(L)}{\subset\mathcal{M}}_Z\subset\mathbb{R}^n$.

\noindent \textbf{Effective homeomorphisms.}  Although the actual neural transformations are from one layer to the next,  between any two layers $l$ and $l'$ is an effective compositional homeomorphism   $\phi^{(\tilde{c})}\defeq \phi^{(\{l,l'\})} : X^{(l)}\ \rightarrow X^{(l')}$. The LIS constraint can be imposed between {\em any} pair of $l$ and $l'$ to constrain $\phi^{(\tilde{c})}$ for all $\tilde{c}\in \tilde{C}_2$ and eventually on the overall $\Phi$.

\noindent \textbf{Evolution in graph structure.} The graphs, as represented by $\mathcal{G}(X^{(l)},D^{(l)},\mathcal{N}^{(l)})$, are evolving from layer to layer as the learning continues except for the input layer. The neighborhood structures are allowed to change from $l$ to $l'$ for admissible geometric deformations caused by the nonlinearity of $\phi^{(\{l,l'\})} $. To account for such changes across-layer, we define the set union $\mathcal{C}_2^{(\tilde{c})} = \mathcal{C}_2^{(\{l,l'\})} =\mathcal{C}_2^{(l)}\cup\mathcal{C}_2^{(l')}$ as the set of pair-wise cliques to be considered in formulating losses for $\phi^{(\{l,l'\})}$ for $\tilde{c}=\{l,l'\}\in \tilde{C}_2$.

\section{ Markov-Lipschitz Neural Networks}

\subsection{Local isometry and Lipschitz continuity}

The core of MLDL is the imposition of the prior LIS constraint across-layers  (as orange-colored arcs and dashed lines in Fig.~\ref{fig:MLDL}). MLDL requires that the homeomorphism $\Phi : X\rightarrow Z$ should satisfy the LIS constraint across layers; that is, the distances (or some other metrics) be preserved locally, $d_X{(x}_i,x_j)=d_Z(\Phi(x_i,\Phi{(x}_j))$ for $j\in\mathcal{N}_{i}$, as far as possible, so as to optimize homeomorphic regularity between metric spaces at different network layers. Such a $\Phi$ can be learned by minimizing the following energy functional
\begin{align}
	\mathcal{L}_{LIS}\!(\Phi\!)=\sum_{i\in\mathcal{S}}\sum_{j\in\mathcal{N}_{i}}\left|d_X\!(x_i\!,x_j\!)\!-d_Z(\Phi(x_i),\Phi(x_j))\right|
\label{equ:math-isometry}.
\end{align}
$\mathcal{L}_{LIS}(\Phi)$ measures the overall geometric distortion and homeomorphic (ir)regularity of the neural transformation $\Phi$ and reaches the lower bound of $0$ when the local isometry constraint is strictly satisfied. $\Phi$ is said to be {\em locally Lipschitz} if for all $j\in\mathcal{N}_{i}$ there exists $K>0$ with
\begin{align*}
d_Z(\Phi(x_i),\Phi(x_j))\ \le\ K \ d_X{(x}_i,x_j).
\end{align*}
This requires that $\Phi$ does not "collapsed", i.e. $\Phi(x_i)\neq \Phi(x_j)$ for $x_i\neq x_j$.  A Lipschitz mapping or homeomorphism with a smaller $K$ tends to generalize better \cite{Bartlett-NIPS-2017,Anil-ICML-2018}, be more robust to adversarial attacks \cite{Weng-ICLR-2019,Cohen-ICML-2019}, and more stable in generative model learning \cite{Zhou-Lip-GAN-ICML2019,Qi-LS-GAN-IJCV2020}. A mapping is {\em locally bi-Lipschitz} if for all $j\in\mathcal{N}_{i}$ there exists $K\geq 1$ with
\begin{align}
d_X{(x}_i,x_j) \!/ K\le d_Z(\Phi\left(x_i\right)\!,\Phi{(x}_j)\!)\le K d_X{(x}_i,x_j) 
\label{equ:math-bi-Lipschitz}.
\end{align}
The best possible bi-Lipschitz constant is $K=1$ and the closer to 1 the better. A good bi-Lipschitz homeomorphism would not only well preserve the local geometric structure but also improve the stability and robustness of resulting MLDL neural networks, as will be validated by our extensive experiments.

\subsection{Markov-Lipschitz encoder (ML-Enc)}

\textbf{ML-Enc for Manifold Learning.} The ML-Enc, unlike the other AEs, can learn an NLDR transformation from the input to latent code by using the LIS constraint only without the need for a reconstruction loss. It consists of a LIS loss plus a transit, auxiliary push-away loss
\begin{align*}
\mathcal{L}_{Enc}(W)=\mathcal{L}_{LIS}(W)+\mu\ \mathcal{L}_{push}(W).
\end{align*} 
in which $\mu$ starts from a positive value at the beginning of manifold learning so that an auxiliary term is effective and gradually decreases to $0$ so that only the target $\mathcal{L}_{LIS}$ takes effect finally. 

\noindent \textbf{LIS loss.} The prior LIS constraint enforces the classic isometry, that of Equ.~(\ref{equ:math-isometry})  and (\ref{equ:math-bi-Lipschitz}), across layers, as a key prior on the manifold learning and NLDR task. It is encoded into cross-layer super-clique potentials which are then summed over all $\tilde{c}=\{l,l^\prime\}\in\widetilde{\mathcal{C}}_2$ into the energy function of an MRF-Gibbs  distribution $U(\widetilde{X})= \sum_{\tilde{c} \in \widetilde{\mathcal{C}}_2 }V_{\tilde{c}}(\widetilde{X})$. The clique potentials due to the LIS constraint are defined as
\begin{align*}
V_{LIS}({\widetilde{X}}\ | \ c,\tilde{c})=\left|d_l(x_i^{\left(l\right)},x_j^{(l)})-d_{l'}(x_i^{(l')},x_j^{(l')})\right|.
\end{align*}
where $\tilde{c}=\{l,l'\}\in\widetilde{\mathcal{C}}_2$ are between-layer super-cliques and ${c=\{i,j\}\in\mathcal{C}}_2^{(\tilde{c})}$ are between-sample cliques. Note that  $X^{(l=0)}=X$ is the given input data and is fixed whereas $X^{(l)}$ for $l>0$ are part of the solution which can be rewritten as $Z^{(l)}\defeq X^{(l)}$. $\mathcal{L}_{LIS}(W)$ actually imposes constraint on the solution $Z^{(l)}$. It corresponds to the energy function in the prior MRF-Gibbs distribution, regardless of the input data $X^{(0)}=X$. 

Summing up the potentials gives rise to the energy function
\begin{align}
\mathcal{L}_{LIS}(W)=
\sum_{\tilde{c}\in\widetilde{\mathcal{C}}_2}\alpha^{(\tilde{c})}\sum_{c\in\mathcal{C}^{(\tilde{c})}_2}V_{LIS}({\widetilde{X}}\ | \ c,\tilde{c}) , 
\label{equ:loss-lis}
\end{align}
where $\mathcal{C}^{(\tilde{c})}_2=\mathcal{C}^{(l)}_2 \cup \mathcal{C}^{(l')}_2$ is the union of the two pairwise clique sets as defined before and $\alpha^{(\tilde{c})}$ are weights. Here the LIS loss $\mathcal{L}_{LIS}(W)$, corresponding to Equ.~(\ref{equ:math-isometry}),  is expressed as a function of $W$ because given the network architecture, $\Phi=\Phi(W)$. The weights $\alpha^{(\tilde{c})}$ determine how the LIS constraint should be imposed across $\tilde{c}$. Wishing that as the layer goes closer and closer to the deepest latent layer $L$, the Euclidean metric makes more and more sense, we will evaluate three schemes: decreasing, increasing and constant as the link goes deeper for a scheme of linking between the input layer and each subsequent ML-Enc layer.  

\noindent \textbf{Push-away loss.} The push-away loss is defined as
\begin{align*}
\mathcal{L}_{push}(W)=-\sum_{\tilde{c}\in\widetilde{\mathcal{C}}_2}
\sum_{c\not \in \mathcal{C}_2^{(l)}}\pi[d_{l'}(x_i^{(l')},x_j^{(l')})<B] \  d_{l'}(x_i^{(l')},x_j^{(l')}), 
\end{align*}
in which $\pi[\cdot]\in\{0,1\}$ is the indicator function and $B$ is a bound. This term is aimed to help "unfold" nonlinear manifolds, by exerting a spring force to push away from each other those pairs $(i,j)$ which are non-neighbors at layer $l$ but nearby (distance smaller than $B$) at layer $l'$.

\subsection{Markov-Lipschitz AutoEncoder (ML-AE)}
 
The ML-AE has two purposes: (1) helping further regularize ML-Enc based manifold learning with the ML-Dec and reconstruction losses, and (2) enabling manifold data generation of the learned manifold. The ML-AE is constructed by appending the ML decoder (ML-Dec) to the ML-Enc, implementing the inverse of the ML-Enc, with reconstruction losses imposed. The ML-AE is symmetric to the ML-Enc in its structure (see Fig.~\ref{fig:MLDL}). Nonetheless,  an asymmetric decoder is also acceptable.

The LIS loss for the ML-Dec can be defined in a similar way to Equ.~(\ref{equ:loss-lis}). The LIS constraint may also be imposed between the corresponding layers of the ML-Enc and ML-Dec.  The {\bf total reconstruction loss} is the sum of individual ones between the corresponding layers (shown as dashed lines in Fig.~1)
\begin{align}
\mathcal{L}_{rec}(W)=\sum_{l=0}^{L-1} \gamma_l \sum_{i=1}^{M}\parallel{x_i}^{\left(l\right)}-{\hat{x}}_i^{\left(l\right)}\parallel^2,
\end{align}
where $\gamma_l$ are weights. The {\bf total ML-AE loss} is then
\begin{align*}
\mathcal{L}_{AE}(W)=\mathcal{L}_{LIS}(W)+ \mathcal{L}_{rec}(W)+\mu\ \mathcal{L}_{push}(W).
\end{align*}
Once trained, the ML decoder can be used to generate new data of the learned manifold. 


\section{Experiments}

The purpose of the experiments is to evaluate the ML-Enc and ML-AE in their ability to preserve local geometry of manifolds and to achieve good stability and robustness in terms of relevant evaluation metrics. While this section presents numerical evaluation of major comparisons, visualization and more numerical results can be found in Appendix.

\noindent \textbf{Four datasets} are used: (i) Swiss Roll  (3-D) and (ii) S-Curve (3-D) generated by the sklearn library \cite{scikit-learn}, (iii)  MNIST (784-D), and (vi) Spheres (101-D) \cite{Moor19Topological}. \textbf{Seven methods for manifold learning} are compared: ML-Enc (ours),  HLLE \cite{donoho2003hessian}, MLLE  and LTSA\cite{zhang2007mlle}, ISOMAP \cite{Tenenbaum-science-00}, LLE \cite{Roweis-science-00} and t-SNE \cite{maaten2008visualizing}; \textbf{Four autoencoder methods} are compared for manifold learning, reconstruction and manifold data generation: ML-AE (ours), AE \cite{hinton2006reducing}, VAE \cite{kingma2013auto}, and TopoAE \cite{Moor19Topological}. 

\textbf{The Euclidean distance metric} is used for all layers ($d_l$) and is normalized to $d'_l=d_l/\sqrt{n_l}$ by the dimensionality $n_l$. The $k$-NN scheme (or $r$-ball ) is used to define {\bf neighborhood systems}. The learning rate is set to 0.001, and the batch size is set to the number of samples. LeakyReLU is used as the \textbf{activation  function}. \textbf{Continuation}: $\mu$ starts from an initial value at epoch 500 and decreases linearly to the minimum 0 at epoch 1000.

\noindent \textbf{Hyper parameters.}
For Swiss Roll and S-Curve, the network structure is [784, 100, 100, 100, 3, 2],  $r=0.23$ (working with range of 0.2~0.3), $B=3$, $\gamma_l=1$ ($l \in \{1,2,\cdots, L\}$), $\alpha^{(0,L)}=1$, and continuation $\mu=0.2\to 0$. 
For MNIST, the network structure is [784,  1000, 500, 250, 100, 2], $K=5$ (in the neighborhood system), $B=2$, $\gamma_l=200$ ($l \in \{1,2,\cdots, L\}$), $\alpha^{(0,L)}=1$, and $\mu=1 \to 0$.
For Spheres5500+5500, the Network structure is [101,  50, 25, 2], $K=15$, $B=3$, $\gamma_l=0$ ($l \in \{1,2,\cdots, L\}$), $\alpha^{(0,L)}=1$, and $\mu=0$. For Spheres10000, the network structure is [101,  90, 80, 70, 2]. Other hyper parameters are the same as Spheres5500+5500.  The implementation is based on the PyTorch library running on Ubuntu 18.04 on NVIDIA v100 GPU. The code is available at \textcolor{red}{https://github.com/westlake-cairi/Markov-Lipschitz-Deep-Learning}


\noindent \textbf{Evaluation metrics} include the following.
(1) The number of successes (\textbf{\#Succ}) is the number of successes (in unfolding the manifold) out of 10 solutions from random initialization (with random seed in $\{0,1,\cdots,9\}$). 
(2) Local KL divergence (\textbf{L-KL}) measures the difference between distributions of local distances in the input and latent spaces. 
(3) Averaged relative rank change (\textbf{ARRC}), (4) Trustworthiness (\textbf{Trust}) and (4') Continuity (\textbf{Cont}) measure how well neighboring relationships are preserved between two spaces (layers).
(5) Locally geometric distortion (\textbf{LGD}) measures how much corresponding distances between neighboring points differ in two metric spaces. 
(6) Mean projection error (\textbf{MPE}) measures the "coplanarity" of 2D embeddings in a high-D space (in the following, the 3D layer before the 2D latent layer). 
(7) Minimum (\textbf{$K$-Min}) and  Maximum (\textbf{$K$-Max}) are for local bi-Lipschitz constant values  Equ.~(\ref{equ:math-bi-Lipschitz}) of computed for all neighborhoods. 
(8) Mean reconstruction error (\textbf{MRE}) measures the difference between the input and output of autoencoders. 
Of the above, the MPE (or "coplanarity") and $K$-Min and $K$-Max are used for the first time as evaluation metrics for manifold learning. Their exact definitions are given in {Appendix A.1}. Every set of experiments is run 10 times with the 10 random seeds, and the results are averaged into the final performance metric.  When a run is unsuccessful, the numerical averages are not very meaningful so the numbers will be shown in \textcolor[rgb]{0.5,0.5,0.5}{gray color} in the following tables.

\begin{table*}[!htb]
	\centering
\caption{Comparison of embedding quality for Swiss Roll (800 points)}
\begin{tabular}{llllllllll}
\toprule
       & \#Succ      & L-KL  & ARRC   & Trust   & Cont    & LGD     & $K$-Min     & $K$-Max    & MPE  \\ \midrule
ML-Enc & \textbf{10} &\textbf{0.0184}   & \textbf{0.000414}   & \textbf{0.9999}  & \textbf{0.9985}  & \textbf{0.00385}                              & \textbf{1.00}                             & \textbf{2.14}                        &\textbf{0.0718}   \\
MLLE   & 6           & \textcolor[rgb]{0.7,0.7,0.7}{0.1251} & \textcolor[rgb]{0.7,0.7,0.7}{0.030702}    & \textcolor[rgb]{0.7,0.7,0.7}{0.9455} & \textcolor[rgb]{0.7,0.7,0.7}{0.9844}          & \textcolor[rgb]{0.7,0.7,0.7}{0.04534}         & \textcolor[rgb]{0.7,0.7,0.7}{7.37}        & \textcolor[rgb]{0.7,0.7,0.7}{238.74} &\textcolor[rgb]{0.7,0.7,0.7}{0.1709}           \\
HLLE   & 6           & \textcolor[rgb]{0.7,0.7,0.7}{0.1297} & \textcolor[rgb]{0.7,0.7,0.7}{0.034619} & \textcolor[rgb]{0.7,0.7,0.7}{0.9388}   & \textcolor[rgb]{0.7,0.7,0.7}{0.9859}          & \textcolor[rgb]{0.7,0.7,0.7}{0.04542}         & \textcolor[rgb]{0.7,0.7,0.7}{7.44}        & \textcolor[rgb]{0.7,0.7,0.7}{218.38} & \textcolor[rgb]{0.7,0.7,0.7}{0.0978}           \\
LTSA   & 6           & \textcolor[rgb]{0.7,0.7,0.7}{0.1296} & \textcolor[rgb]{0.7,0.7,0.7}{0.034933} & \textcolor[rgb]{0.7,0.7,0.7}{0.9385}   & \textcolor[rgb]{0.7,0.7,0.7}{0.9859}          & \textcolor[rgb]{0.7,0.7,0.7}{0.04542}         & \textcolor[rgb]{0.7,0.7,0.7}{7.44}        & \textcolor[rgb]{0.7,0.7,0.7}{215.93} & \textcolor[rgb]{0.7,0.7,0.7}{0.0964}           \\
ISOMAP & 6           & \textcolor[rgb]{0.7,0.7,0.7}{0.0234} & \textcolor[rgb]{0.7,0.7,0.7}{0.009650}  & \textcolor[rgb]{0.7,0.7,0.7}{0.9827}  & \textcolor[rgb]{0.7,0.7,0.7}{0.9950}          & \textcolor[rgb]{0.7,0.7,0.7}{0.02376}         & \textcolor[rgb]{0.7,0.7,0.7}{1.11}        & \textcolor[rgb]{0.7,0.7,0.7}{34.35}  & \textcolor[rgb]{0.7,0.7,0.7}{0.0429}           \\
t-SNE  & 0           & \textcolor[rgb]{0.7,0.7,0.7}{0.0450} & \textcolor[rgb]{0.7,0.7,0.7}{0.006108}  & \textcolor[rgb]{0.7,0.7,0.7}{0.9987}  & \textcolor[rgb]{0.7,0.7,0.7}{0.9843}          & \textcolor[rgb]{0.7,0.7,0.7}{3.40665}         & \textcolor[rgb]{0.7,0.7,0.7}{11.1}        & \textcolor[rgb]{0.7,0.7,0.7}{1097.62}& \textcolor[rgb]{0.7,0.7,0.7}{0.1071}           \\
LLE    & 0           & \textcolor[rgb]{0.7,0.7,0.7}{0.1775} & \textcolor[rgb]{0.7,0.7,0.7}{0.014249}  & \textcolor[rgb]{0.7,0.7,0.7}{0.9753}  & \textcolor[rgb]{0.7,0.7,0.7}{0.9895}          & \textcolor[rgb]{0.7,0.7,0.7}{0.04671}         & \textcolor[rgb]{0.7,0.7,0.7}{6.17}        & \textcolor[rgb]{0.7,0.7,0.7}{451.58} & \textcolor[rgb]{0.7,0.7,0.7}{0.1400}           \\ 
TopoAE  & 0           & \textcolor[rgb]{0.7,0.7,0.7}{0.0349} & \textcolor[rgb]{0.7,0.7,0.7}{0.022174}  & \textcolor[rgb]{0.7,0.7,0.7}{0.9661}  & \textcolor[rgb]{0.7,0.7,0.7}{0.9884}          & \textcolor[rgb]{0.7,0.7,0.7}{0.13294}         & \textcolor[rgb]{0.7,0.7,0.7}{1.27}        & \textcolor[rgb]{0.7,0.7,0.7}{189.95}& \textcolor[rgb]{0.7,0.7,0.7}{0.1307}               \\ \bottomrule
\end{tabular}
\label{tbl:swiss800} 
\end{table*}

\begin{center}
	\begin{table*}[!htb]
	\centering
		\caption{\label{tbl:pointsandnoise}Times of successes in 10 runs with varying numbers of samples and noise levels}
		\begin{tabular}{lllllllllllll}
\toprule
            & \multicolumn{5}{c}{\#Samples} &  & \multicolumn{6}{c}{Noise level}\\
            \cmidrule{2-6} \cmidrule{8-13} 
            & 700           & 800           & 1000          & 1500          & 2000          &  & 0.05       & 0.10          & 0.15          & 0.20          & 0.25      & 0.30          \\ \midrule
ML-Enc      & \textbf{10}   & \textbf{10}   & \textbf{10}   & \textbf{10}   & \textbf{10}   &  & \textbf{10}& \textbf{10}   & \textbf{10}   & \textbf{9}    & \textbf{8}& \textbf{8}    \\
MLLE        & 2             & 7             & 7             & \textbf{10}   & \textbf{10}   &  & 7          & 7             & 6             & 6             & 5         & 4             \\
HLLE        & 2             & 6             & 7             & \textbf{10}   & \textbf{10}   &  & 7          & 7             & 6             & 6             & 5         & 3             \\
LTSA        & 2             & 6             & 7             & \textbf{10}   & \textbf{10}   &  & 7          & 7             & 6             & 6             & 5         & 3             \\
ISOMAP      & 2             & 6             & 7             & \textbf{10}   & \textbf{10}   &  & 7          & 7             & 6             & 6             & 5         & 4             \\
\bottomrule
		\end{tabular}
	\end{table*}
\end{center}

\subsection{ML-Enc for manifold learning and NLDR}


\textbf{NLDR quality.} Table~\ref{tbl:swiss800} compares the ML-Enc with 7 other methods (TopoAE also included) in 9 evaluation metrics, using the Swiss Roll (800 points) manifold data (the higher \#Succ, Trust and Cont are, the better; the lower the other metrics, the better). Results with the S-Curve are given in {Appendix A.2}. While the MPE is calculated at layer $4$ (3D), the other metrics are calculated between the input and the latent layer. The results demonstrate that the ML-Enc outperforms all the other methods in all the evaluation metrics, particularly significant in terms of the isometry (LGD, ARRC, Trust and Cont) and Lipschitz ($K$-Min and $K$-Max) related metrics. 

\noindent \textbf{Robustness to sparsity and noise.} Table~\ref{tbl:pointsandnoise} evaluates the success rates  of 5 manifold learning methods (t-SNE and LLE not included because they had zero success) in their ability to unfold the manifold and robustness to varying numbers of samples (700, 800, 1000, 1500, 2000) and  standard deviation of noise $\sigma=\ 0.05,\ 0.10,\ 0.15,\ 0.20 ,\ 0.25 ,\ 0.30 $. The corresponding evaluation metrics are provided in {Appendix A.3}. These two sets of experiments demonstrate that the ML-Enc achieves the highest success rate, the best performance metrics, and the best robustness to data sparsity and noise.

\noindent \textbf{Generalization to unseen data.} The ML-Enc trained with 800 points can generalize well to unfold unseen samples of the learned manifold. The test is done as follows: First, a set of 8000 points of the Swiss Roll manifold are generated; the data set is modified by removing, from the generated 8000 points of the manifold, the shape of a diamonds, square, pentagram, or five-ring, respectively, creating 4 test sets. Each point of a test set is transformed independently by the trained ML-Enc to obtain an embedding ( shown in Appendix A.4). We can see that the unseen manifold data sets are well unfolded by the ML-Enc and the removed shapes are kept very well, illustrating that the learned ML-Enc has a good ability to generalize to unseen data. Since LLE-based, LTSA, and ISOMAP algorithms do not possess such a generalization ability, the ML-Enc is compared with the encoder parts of the AE based algorithms. Unfortunately, AE and VAE failed altogether for the Swiss Roll data sets.

\subsection{ML-AE for manifold generation}


{\bf Manifold reconstruction.} This set of experiments compare the ML-AE with AE \cite{hinton2006reducing}, VAE \cite{kingma2013auto}, and TopoAE \cite{Moor19Topological}. Table~\ref{tbl:reconstrction} compares the 9 quality metrics (each value being calculated as the averages for the 10 runs) for the 4 autoencoders with the Swiss Roll (800 points) data. The MPE is the average of the MPE's at layers $4$ and $4'$ and the other metrics are calculated between the input and output layers of the AE's. While the other 3 autoencoders fail to unfold the manifold data sets ({\it hence their metrics do not make much sense}), the ML-AE produces good quality results especially in terms of the isometry and Lipschitz related metrics. The resulting metrics also suggest that $K$-max could be used as an indicator of success/failure in manifold unfolding.

\begin{center}
\begin{table*}[!htb]
	\centering
    \caption{Performance metrics for the ML-AE with Swiss Roll (800 points) data \label{tbl:reconstrction}}
	\begin{tabular}{llllllllll}
		\toprule
		            & \#Succ            & L-KL                                  & ARRC                                     & Cont                                    & LGD                                   & $K$-min                               & $K$-max                               & MPE               & MRE       \\
		\midrule
		ML-AE       & \textbf{10}       & \textbf{0.00165}                      & \textbf{0.00070}                        & \textbf{0.9998}                         & \textbf{0.00514}                      & \textbf{1.01}                         & \textbf{2.54}                         & \textbf{0.04309}           & \textbf{0.01846}  \\
		AE          & 0                 & \textcolor[rgb]{0.7,0.7,0.7}{0.11537} & \textcolor[rgb]{0.7,0.7,0.7}{0.13589}   & \textcolor[rgb]{0.7,0.7,0.7}{0.9339}    & \textcolor[rgb]{0.7,0.7,0.7}{0.03069} & \textcolor[rgb]{0.7,0.7,0.7}{1.82}    & \textcolor[rgb]{0.7,0.7,0.7}{5985.74} & \textcolor[rgb]{0.7,0.7,0.7}{0.01519} & \textcolor[rgb]{0.7,0.7,0.7}{0.40685}     \\
		VAE         & 0                 & \textcolor[rgb]{0.7,0.7,0.7}{0.23253} & \textcolor[rgb]{0.7,0.7,0.7}{0.49784}   & \textcolor[rgb]{0.7,0.7,0.7}{0.5039}    & \textcolor[rgb]{0.7,0.7,0.7}{0.04000} & \textcolor[rgb]{0.7,0.7,0.7}{1.49}    & \textcolor[rgb]{0.7,0.7,0.7}{5290.55} & \textcolor[rgb]{0.7,0.7,0.7}{0.01977} & \textcolor[rgb]{0.7,0.7,0.7}{0.78104}     \\ 
		TopoAE      & 0                 & \textcolor[rgb]{0.7,0.7,0.7}{0.05793} & \textcolor[rgb]{0.7,0.7,0.7}{0.04891}   & \textcolor[rgb]{0.7,0.7,0.7}{0.9762}    & \textcolor[rgb]{0.7,0.7,0.7}{0.09651} & \textcolor[rgb]{0.7,0.7,0.7}{1.10}    & \textcolor[rgb]{0.7,0.7,0.7}{228.11}  & \textcolor[rgb]{0.7,0.7,0.7}{0.12049} & \textcolor[rgb]{0.7,0.7,0.7}{0.56013}     \\
		\bottomrule
\end{tabular}
\end{table*}
\end{center}

\noindent \textbf{Manifold data generation.}
The ML-Dec part of the trained ML-AE can be used to generate new data of the learned manifold, mapping from random samples in the latent space to points in the ambient input space. The generated manifold data points, shown in Appendix A.5, are well-behaved due to the bi-Lipschitz continuity. It also suggests that it is possible to construct invertible, bijective mappings between spaces of different dimensionalities using the proposed MLDL method.

\subsection{Results with high-dimensional datasets}

Having evaluated toy datasets with perceivable structure, the following presents the results with the {\rm MNIST} dataset in 784-D and the {\bf Spheres} dataset in 101-D spaces, obtained by using ML-Enc and ML-AE  in comparison with others, shown in Table~\ref{tbl:mnist+sphere} (and Figures.~A1, A2 and A3 in Appendix A.2). For the {\bf MNIST} dataset of 10 digits, a subset of 8000 points are randomly chosen for training and another subset of 8000 points for testing without overlapping. The ML-Enc is used for NLDR by manifold learning. The results are shown in the first parts of Table~\ref{tbl:mnist+sphere}.

\begin{center}
\begin{table*}[!htb]
\centering
\caption{Performance metrics on MNIST(top) and Spheres5500+5500(bottom)}
\label{tbl:mnist+sphere}
\begin{tabular}{l|l|ccccccc}
\toprule
&\textbf{} & \textbf{L-KL}     & \textbf{ARRC}   & \textbf{Trust}   & \textbf{Cont}    & \textbf{K-min}     & \textbf{K-max}    & \textbf{LGD}      \\
\midrule
\multirow{5}{*}{MNIST} 				&ISOMAP 							& 0.428             & \textbf{0.003} &  0.994           & 0.993            & 6.395              & 23.62             & 2.629             \\
									&MLLE                                & 0.227             & 0.392          &  0.609           & 0.554            & 1.773              & 7.4 E+11          & 0.180             \\
									&t-SNE                               & \textbf{0.120}    & 0.005          &  0.994           & 0.988            & 1.137              & 14385.0           & 0.192             \\
									&TopoAE                              & 0.174             & 0.006          &  \textbf{0.999}  & 0.993            & 3.200              & 7.012             & 0.298             \\
									&ML-Enc (ours)                       & 0.324             & 0.004          &  0.991           & \textbf{0.998}   & \textbf{1.006}     & \textbf{1.273}    & \textbf{0.006}    \\ 
\toprule

\multirow{5}{*}{Spheres5500+5500}   &ISOMAP                              & 0.082             & 0.091          &  0.881           & 0.933             & 1.241             & 194.5             & 0.251             \\
									&MLLE                                & 0.233             & 0.095          &  0.880           & 0.924             & 1.693             & 862.1             & 0.041             \\
									&t-SNE                               & 0.079             & \textbf{0.046} &  \textbf{0.924}  & \textbf{0.968}    & 1.526             & 690.5             & 6.814             \\
									&TopoAE                              & 0.103             & 0.091          &  0.883           & 0.940             & 1.195             & 77.5              & 0.507             \\
									&ML-Enc (ours)                       & \textbf{0.007}    & 0.095          &  0.879           & 0.931             & \textbf{1.023}    & \textbf{40.9}     &\textbf{0.021}     \\ 
\bottomrule
\end{tabular}
\end{table*}
\end{center}

\begin{center}
\begin{table*}[!htb]
	\centering
\caption{Ablation of three losses imposed between layers 0 and 5 for ML-Enc and between layers 0 and 0' for ML-AE}
\label{tbl:ablation-AE}
\begin{tabular}{lccccccccc}
\toprule
            & \#Succ         & L-KL                                             & ARRC                                   & Cont                                  & LGD                                       & $K$-Min                               & $K$-Max         & MPE               & MRE               \\ \midrule
		AB  & \textbf{10} & 0.01842                                 & \textbf{0.00041}                      & \textbf{0.9998}                       & \textbf{0.00385} & \textbf{1.00} & \textbf{2.14} & 0.04309  & N/A    \\
		A   & 0           & \textcolor[rgb]{0.7,0.7,0.7}{0.06257}   & \textcolor[rgb]{0.7,0.7,0.7}{0.09208} & \textcolor[rgb]{0.7,0.7,0.7}{0.9810}  & \textcolor[rgb]{0.7,0.7,0.7}{0.00434} & \textcolor[rgb]{0.7,0.7,0.7}{1.00} & \textcolor[rgb]{0.7,0.7,0.7}{5.76} & \textcolor[rgb]{0.7,0.7,0.7}{0.00574} & N/A \\
		B   & 0           & \textcolor[rgb]{0.7,0.7,0.7}{0.05213}   & \textcolor[rgb]{0.7,0.7,0.7}{0.09407} & \textcolor[rgb]{0.7,0.7,0.7}{0.9806} & \textcolor[rgb]{0.7,0.7,0.7}{27.76862} & \textcolor[rgb]{0.7,0.7,0.7}{177.41} & \textcolor[rgb]{0.7,0.7,0.7}{1089.54} & \textcolor[rgb]{0.7,0.7,0.7}{0.00374}   & N/A \\
 \midrule
		AB          & \textbf{10}    & \textbf{0.00165}                                 & \textbf{0.00070}                      & \textbf{0.9998}                       & \textbf{0.00514}                          & \textbf{1.01}                         & \textbf{2.54} & 0.04309          & \textbf{0.01846}  \\
A           & 0              & \textcolor[rgb]{0.7,0.7,0.7}{0.07579}           & \textcolor[rgb]{0.7,0.7,0.7}{0.09127}  & \textcolor[rgb]{0.7,0.7,0.7}{0.9810}  & \textcolor[rgb]{0.7,0.7,0.7}{0.02776}     & \textcolor[rgb]{0.7,0.7,0.7}{1.05}    & \textcolor[rgb]{0.7,0.7,0.7}{11.34}         & \textcolor[rgb]{0.7,0.7,0.7}{0.00574}           & \textcolor[rgb]{0.7,0.7,0.7}{0.24499}           \\
B           & 0              & \textcolor[rgb]{0.7,0.7,0.7}{0.06916}           & \textcolor[rgb]{0.7,0.7,0.7}{0.08457}  & \textcolor[rgb]{0.7,0.7,0.7}{0.9806}  & \textcolor[rgb]{0.7,0.7,0.7}{0.01896}     & \textcolor[rgb]{0.7,0.7,0.7}{1.05}    & \textcolor[rgb]{0.7,0.7,0.7}{125.69}        & \textcolor[rgb]{0.7,0.7,0.7}{0.00373}           & \textcolor[rgb]{0.7,0.7,0.7}{0.25894}           \\
\bottomrule
\end{tabular}
\end{table*}
\end{center}

The {\bf Spheres10000} dataset, proposed by TopoAE \cite{Moor19Topological}, is composed of 1 big sphere enclosing 10 small ones in 101-D space. {\bf Spheres5500+5500} differs in that its big sphere consists of only 500 samples  (whereas that in Spheres10000 has 5000) -- the data is so sparse that the smallest within-sphere distance on the big sphere can be larger than that between the big sphere and some small ones.  The results are shown in the lower parts of Table~\ref{tbl:mnist+sphere} and in Appendix A.2. 
From Table~\ref{tbl:mnist+sphere} (and Appendix A.2), we can see the following: 
\begin{enumerate}
    \item For the MNIST dataset, the ML-Enc achieves overall the best in terms of the performance metric values and possesses an ability to generalize to unseen test data whereas the other compared methods cannot. Visualization-wise, however, t-SNE delivers the most appealing result.
    \item For the Spheres dataset, both the ML-AE and the ML-Enc (without a decoder) perform better than the SOTA TopoAE and the ML-AE generalizes better than the ML-Enc.
    \item the ML-AE and the ML-Enc handle sparsity better and generalize better than the others.
\end{enumerate}
Overall, the MLDL framework has demonstrated its  superiority over the compared methods in terms of Lipschitz constant and isometry-related properties, realizing its promise.

\subsection{Ablation study}

This evaluates effects of the two loss terms in the ML-AE on the 9 performance metrics, with the Swiss Roll (800 points) data: (A) the LIS loss $\mathcal{L}_{LIS}$ and  (B) the push-away loss $\mathcal{L}_{push}$. Table~\ref{tbl:ablation-AE} shows the results when the LIS loss is applied between layers 0 and and between layers 0 and 0'. The conclusion is: (1) the LIS loss (A) is the most important for achieving the results of excellence; (2) the push-away term (B), which is applied with decreasing and diminishing  weight on convergence, helps unfold manifolds especially with challenging input. Overall, the "AB" is the best combination to make the algorithms work. See more  in {Appendix~A.6}.


\section{Conclusions}

The proposed MLDL framework imposes the cross-layer LIS prior constraint to optimize neural transformations in terms of local geometry preservation and local bi-Lipschitz constants. Extensive experiments with manifold learning, reconstruction, and generation consistently demonstrate that the MLDL networks (ML-Enc and ML-AE) well preserve local geometry for manifold learning and data generation and achieve excellent local bi-Lipschitz constants, advancing deep learning techniques. The main ideas of MLDL are general and effective enough and are potentially applicable to a wide range of neural networks for improving representation learning, data generation, and network stability and robustness. Future work includes the following:  (1) While
LIS preserves the Euclidean distance, nonlinear forms of metric-preserving will be
explored to allow more flexibility for more complicated nonlinear tasks; (2) developing invertible, bijective neural network mappings using the LIS constraint with bi-Lipschitz continuity; (3) extending the unsupervised version of MLDL to self-supervise, semi-supervised and supervised tasks; (4) further formulating MLDL so that cross-layer link hyperparameters $\alpha$ become part of learnable parameters.

\section*{Acknowledgments} 

We would like to acknowledge funding support from the Westlake University and Bright Dream Joint Institute for Intelligent Robotics and thank Zicheng Liu, Zhangyang Gao, Haitao Lin, and  Yiming Qiao for their assistance in processing experimental results. Thanks are also given to Zhiming Zhou for his helpful comments and suggestions for improving the manuscript.

\bibliographystyle{plain} 
\bibliography{vision,MLDL}

\begin{thebibliography}{10}

\bibitem{Anil-ICML-2018}
Cem Anil, James Lucas, and Roger Grosse.
\newblock Sorting out lipschitz function approximation.
\newblock {\em arXiv preprint arXiv:1811.05381}, 2018.

\bibitem{awiszus_markov_2018}
Maren Awiszus and Bodo Rosenhahn.
\newblock Markov chain neural networks.
\newblock {\em CoRR}, abs/1805.00784, 2018.

\bibitem{Bartlett-NIPS-2017}
Peter {Bartlett}, Dylan~J. {Foster}, and Matus {Telgarsky}.
\newblock {Spectrally-normalized margin bounds for neural networks}.
\newblock {\em arXiv e-prints}, page arXiv:1706.08498, June 2017.

\bibitem{Besag-74}
J.~Besag.
\newblock ``{Spatial} interaction and the statistical analysis of lattice
  systems'' (with discussions).
\newblock {\em Journal of the Royal Statistical Society, Series B},
  36:192--236, 1974.

\bibitem{Bronstein-GDL-2017}
Michael~M. {Bronstein}, Joan {Bruna}, Yann {LeCun}, Arthur {Szlam}, and Pierre
  {Vandergheynst}.
\newblock {Geometric Deep Learning: Going beyond Euclidean data}.
\newblock {\em IEEE Signal Processing Magazine}, 34(4):18--42, July 2017.

\bibitem{CHEN-JASA-2009}
Lisha Chen and Andreas Buja.
\newblock Local multidimensional scaling for nonlinear dimension reduction,
  graph drawing, and proximity analysis.
\newblock {\em Journal of the American Statistical Association},
  104(485):209--219, 2009.

\bibitem{Cohen-ICML-2019}
Jeremy~M {Cohen}, Elan {Rosenfeld}, and J.~{Zico Kolter}.
\newblock {Certified Adversarial Robustness via Randomized Smoothing}.
\newblock {\em arXiv e-prints}, page arXiv:1902.02918, February 2019.

\bibitem{donoho2003hessian}
David~L Donoho and Carrie Grimes.
\newblock Hessian eigenmaps: Locally linear embedding techniques for
  high-dimensional data.
\newblock {\em Proceedings of the National Academy of Sciences},
  100(10):5591--5596, 2003.

\bibitem{Fefferman-manifold-2016}
Charles Fefferman, Sanjoy Mitter, and Hariharan Narayanan.
\newblock Testing the manifold hypothesis.
\newblock {\em Journal of American Mathematical Society}, 29(4):983–1049,
  2016.

\bibitem{Gashler-NIPS-2007}
Michael Gashler, Dan Ventura, and Tony Martinez.
\newblock Iterative non-linear dimensionality reduction with manifold
  sculpting.
\newblock In J.~C. Platt, D.~Koller, Y.~Singer, and S.~T. Roweis, editors, {\em
  Advances in Neural Information Processing Systems 20}, pages 513--520, 2008.

\bibitem{Geman-Geman-84}
Stuart Geman and Donald Geman.
\newblock ``{Stochastic} relaxation, {Gibbs} distribution and the {Bayesian}
  restoration of images''.
\newblock {\em IEEE Transactions on Pattern Analysis and Machine Intelligence},
  6(6):721--741, November 1984.

\bibitem{Gulrajani-NIPS-2017}
Ishaan Gulrajani, Faruk Ahmed, Martín Arjovsky, Vincent Dumoulin, and Aaron~C.
  Courville.
\newblock Improved training of wasserstein gans.
\newblock In {\em Advances in Neural Information Processing Systems}, pages
  5767--5777, 2017.

\bibitem{hinton2006reducing}
Geoffrey~E Hinton and Ruslan~R Salakhutdinov.
\newblock Reducing the dimensionality of data with neural networks.
\newblock {\em science}, 313(5786):504--507, 2006.

\bibitem{kingma2013auto}
Diederik~P Kingma and Max Welling.
\newblock Auto-encoding variational bayes.
\newblock {\em arXiv preprint arXiv:1312.6114}, 2013.

\bibitem{Latorre-ICLR-2020}
Fabian Latorre, Paul Rolland, and Volkan Cevher.
\newblock Lipschitz constant estimation of neural networks via sparse
  polynomial optimization.
\newblock In {\em International Conference on Learning Representations}, 2020.

\bibitem{Li-MRF-Book-95}
Stan~Z. Li.
\newblock {\em Markov Random Field Modeling in Computer Vision}.
\newblock Springer, 1995.

\bibitem{maaten2008visualizing}
Laurens van~der Maaten and Geoffrey Hinton.
\newblock Visualizing data using t-sne.
\newblock {\em Journal of machine learning research}, 9(Nov):2579--2605, 2008.

\bibitem{McQueen-NIPS-2016}
James McQueen, Marina Meila, and Dominique Joncas.
\newblock Nearly isometric embedding by relaxation.
\newblock In D.~D. Lee, M.~Sugiyama, U.~V. Luxburg, I.~Guyon, and R.~Garnett,
  editors, {\em Advances in Neural Information Processing Systems 29}, pages
  2631--2639, 2016.

\bibitem{Belkin-Niyogi-02}
Partha~Niyogi Mikhail~Belkin.
\newblock Laplacian eigenmaps for dimensionality reduction and data
  representation.
\newblock {\em Technical Report, University of Chicago}, January 2002.

\bibitem{Miyato-ICLR-2018}
Takeru Miyato, Toshiki Kataoka, Masanori Koyama, and Yuichi Yoshida.
\newblock Spectral normalization for generative adversarial networks.
\newblock In {\em International Conference on Learning Representations}, pages
  5767--5777, 2018.

\bibitem{Moor19Topological}
Michael Moor, Max Horn, Bastian Rieck, and Karsten Borgwardt.
\newblock Topological autoencoders.
\newblock In {\em Proceedings of the 37th International Conference on Machine
  Learning~(ICML)}, Proceedings of Machine Learning Research. PMLR, 2020.

\bibitem{scikit-learn}
F.~Pedregosa, G.~Varoquaux, A.~Gramfort, V.~Michel, B.~Thirion, O.~Grisel,
  M.~Blondel, P.~Prettenhofer, R.~Weiss, V.~Dubourg, J.~Vanderplas, A.~Passos,
  D.~Cournapeau, M.~Brucher, M.~Perrot, and E.~Duchesnay.
\newblock Scikit-learn: Machine learning in {P}ython.
\newblock {\em Journal of Machine Learning Research}, 12:2825--2830, 2011.

\bibitem{Qi-LS-GAN-IJCV2020}
Guo-Jun Qi.
\newblock Loss-sensitive generative adversarial networks on lipschitz
  densities.
\newblock {\em International Journal of Computer Vision}, pages 1--23, 2019.

\bibitem{Roweis-science-00}
Sam~T Roweis and Lawrence~K Saul.
\newblock Nonlinear dimensionality reduction by locally linear embedding.
\newblock {\em science}, 290(5500):2323--2326, 2000.

\bibitem{Tenenbaum-science-00}
Joshua~B Tenenbaum, Vin De~Silva, and John~C Langford.
\newblock A global geometric framework for nonlinear dimensionality reduction.
\newblock {\em science}, 290(5500):2319--2323, 2000.

\bibitem{Virmaux-NIPS-2018}
Aladin Virmaux and Kevin Scaman.
\newblock Lipschitz regularity of deep neural networks: analysis and efficient
  estimation.
\newblock In S.~Bengio, H.~Wallach, H.~Larochelle, K.~Grauman, N.~Cesa-Bianchi,
  and R.~Garnett, editors, {\em Advances in Neural Information Processing
  Systems 31}, pages 3835--3844. NIPS, 2018.

\bibitem{Wasserman-2016}
Larry Wasserman.
\newblock {Topological Data Analysis}.
\newblock {\em arXiv e-prints}, September 2016.

\bibitem{Weng-ICLR-2019}
Tsui-Wei {Weng}, Huan {Zhang}, Pin-Yu {Chen}, Jinfeng {Yi}, Dong {Su}, Yupeng
  {Gao}, Cho-Jui {Hsieh}, and Luca {Daniel}.
\newblock {Evaluating the Robustness of Neural Networks: An Extreme Value
  Theory Approach}.
\newblock {\em arXiv e-prints}, page arXiv:1801.10578, January 2018.

\bibitem{You-graph-NN-2020}
Jiaxuan You, Jure Leskovec, Kaiming He, and Saining~Xie 2.
\newblock {Graph Structure of Neural Networks}.
\newblock {\em arXiv e-prints}, July 2020.

\bibitem{Zhang-Lip-Estimation-NIPS-2018}
Huan Zhang, Tsui-Wei Weng, Pin-Yu Chen, Cho-Jui Hsieh, and Luca Daniel.
\newblock Efficient neural network robustness certification with general
  activation functions.
\newblock In {\em Advances in Neural Information Processing Systems}, pages
  4939--4948, 2018.

\bibitem{zhang2007mlle}
Zhenyue Zhang and Jing Wang.
\newblock Mlle: Modified locally linear embedding using multiple weights.
\newblock In {\em Advances in Neural Information Processing systems}, pages
  1593--1600, 2007.

\bibitem{Zhou-Lip-GAN-ICML2019}
Zhiming Zhou, Jiadong Liang, Yuxuan Song, Lantao Yu, Hongwei Wang, Weinan
  Zhang, Yong Yu, and Zhihua Zhang.
\newblock Lipschitz generative adversarial nets.
\newblock {\em arXiv preprint arXiv:1902.05687}, 2019.

\end{thebibliography}

\setcounter{table}{0} 
\setcounter{figure}{0} 
\renewcommand{\thetable}{A\arabic{table}}
\renewcommand{\thefigure}{A\arabic{figure}}

\section*{Appendix}

\subsection*{A.1 Definitions of performance metrics}

We adopted most of the performance metrics used in TopoAE \cite{Moor19Topological} because ther are suitable for evaluating geometry- and topology-based manifold learning and data generation. We restrict the cross-layer related metrics to those concerning two metric spaces, namely, the input space indexed by $l=0$ and the latent space index by $l'=L$ because the other compared algorithms do not use other cross-layer constrains. The following notations are used in the definitions:
\begin{itemize}
    \item[$d^{(l)}_{i,j}$]: the pairwise distance in space $X^{(l)}$ (i.e. the input space $X$); 
    \item[$d^{(l')}_{i,j}$]: the pairwise distance in space $X^{(l')}$ (i.e. the latent space $Z$);
    \item[$\mathcal{N}_{i,k}^{(l)}$]: the set of indices to the $k$-nearest neighbors ($k$-NN) of  $x^{(l)}_i$;
    \item[$\mathcal{N}_{i,k}^{(l')}$]:  the set of indices to the $k$-NN of  $x^{(l')}_i$;
    \item[$r^{(l)}_{i,j}$]: 
    the closeness rank of $x^{(l)}_j$ in the $k$-NN of $x^{(l)}_i$;
    \item[$r^{(l')}_{i,j}$]: the closeness rank of $x^{(l')}_j$ in the $k$-NN of $x^{(l')}_i$.
\end{itemize}

The evaluation metrics are defined below:

\begin{enumerate}
\setlist[itemize]{leftmargin=*}
\setlist[itemize]{leftmargin=6mm}
    \item[(1)] \textbf{ \#Succ} is the number of times, out of 10 runs each with a random seed, where a manifold learning method has successfully unfolded the 3D manifold (the Swiss Roll or S-Curve) in the input to a 2D planar embedding without twists, tearing, or other defects.

    \item[(2)] \textbf{L-KL} (Local KL divergence) measures the discrepancy between distributions of local distances in two spaces, defined as 
    \begin{align*}
    KL=\sum_{\{i, j\}\in{ \mathcal{C}_2}^{(l,l')}} u^{(l)}_{i,j}  \log \left(\frac{u^{(l)}_{i,j}}{u^{(l')}_{i,j}}\right)
    \end{align*}
where $u^{(l)}_{i,j}$ is the "similarity" (a nonlinear function of the distance) between $i$ and $j$ at layer $l$, defined as
    \begin{align*}
    u^{(l)}_{i,j}= \frac{\exp \left(-\frac{{d^{(l)}_{i,j}}^{2}}{\sigma}\right)}{\sum_{j\in \mathcal{N}_i^{(l)}} \exp \left(-\frac{{d^{(l)}_{i,j}}^{2}}{\sigma}\right)}, 
    \end{align*}
where $\sigma$ is the locality parameter.

    \item[(3)] \textbf{ARRC} (Averaged relative rank change) measures the average of changes in neighbor ranking 
    between two spaces (layers) $l$ and $l'$:
    \begin{align*}
     RRE = \frac{1}{(k_2-k_1+1)}\sum_{k=k_1}^{k_2} \left\{M R^{(l,l')}_k + M R^{(l',l)}_k\right \},
	\end{align*}
	where $k_1$ and $k_2$ are the lower and upper bounds of the $k$-NN, and
    \begin{align*}
    M R^{(l',l)}_k=\frac{1}{H_{k}} \sum_{i=1}^{M} \sum_{j \in \mathcal{N}_{i,k}^{(l)}}\frac{|r^{(l)}_{i,j}-r^{(l')}_{i,j}|}{r^{(l)}_{i,j}},
    \end{align*}
    \begin{align*}
     M R^{(l,l')}_k=\frac{1}{H_{k}} \sum_{i=1}^{M} \sum_{j \in \mathcal{N}_{i,k}^{(l')}}\frac{|r^{(l')}_{i,j}-r^{(l)}_{i,j}|}{r^{(l')}_{i,j})}),
    \end{align*}
in which $H_{k}$ is the normalizing term 
    \begin{align*}H_{k}=M \sum_{k'=1}^{k} \frac{|M-2 k'|}{k'}.\end{align*}
 
    \item[(4)]  \textbf{Trust}  (Trustworthiness)  measures how well the $k$ nearest neighbors of a point are preserved when going from space $X^{(l)}$ to space $X^{(l')}$:
    \begin{align*}
    Trust=\frac{1}{k_2-k_1+1} \sum_{k=k_{1}}^{k_{2}}  \{1-\frac{2}{Mk  (2 M-3 k-1)} \\ \sum_{i=1}^{M} \sum_{j \in \mathcal{N}_{i,k}^{(l')},j \not\in \mathcal{N}_{i,k}^{(l)}}(r^{(l)}_{i,j}-k) \}
    \end{align*}
	where $k_1$ and $k_2$ are the bounds of the number of nearest neighbors, so averaged for different $k$-NN numbers. 
	
	\item[(4')]  \textbf{Cont}  (Continuity)  is asymmetric to \textbf{Trust} (from space $X^{(l')}$ to space $X^{(l)}$):
    \begin{align*}
    Cont=\frac{1}{k_2-k_1+1} \sum_{k=k_{1}}^{k_{2}}  \{1-\frac{2}{Mk  (2 M-3 k-1)} \\ \sum_{i=1}^{M} \sum_{j \in \mathcal{N}_{i,k}^{(l)},j \not\in \mathcal{N}_{i,k}^{(l')}}(r^{(l')}_{i,j}-k) \}
    \end{align*}

    \item[(5)] \textbf{LGD} (Locally geometric distortion) measures how much corresponding distances between neighboring points differ in two metric spaces and is the primary metric for isometry, defined as (with $k_1=4$, and $k_2=10$):
    \begin{align*}
    LGD = \sum_{k=k_{1}}^{k_{2}} \sqrt{\sum_{i}^M \frac{ \sum_{j \in \mathcal{N}_{i,k}^{(l)}} \left(d^{(l)}_{i,j}-d^{(l')}_{i,j}\right)^2 }{  \left(k_{2}-k_{1}+1\right)^2  M(\#\mathcal{N}_i) }}.
    \end{align*}
    where $\#N_i$ is the size of $N_i$ used in MLDL neural network training.
    \item[(6)]  \textbf{MPE} (Mean projection error) measures the "coplanarity" of a set of 3D points $X=\{x_1, \cdots, x_M\}$ (in this work, the 3D layer before the final 2D latent layer).  The least-squares 2D plane in the 3D space is fitted from the 3D points $X$. The 3D points $X$ is projected onto the fitted planes as $P=\{p_1, \cdots, p_M\}$. The MPE is defined as
    \begin{align*}
    MPE=\frac{1}{M} \sum_{i=1}^{M} || x_{i}-p_i ||_2.
    \end{align*}

    \item [(7)] \textbf{$K$-min} and  \textbf{$K$-max}  are the minimum and maximum of the local bi-Lipschitz constant for the homeomorphism between layers $l$ and $l'$, with respect to the given neighborhood system:

	\begin{align*}
		K_{\rm Min}=\min _{i=1}^{M} \max _{j \in \mathcal{N}_{i,k}^{(l)} } K_{i, j}, 
	\end{align*}
	\begin{align*}
		K_{\rm Max}=\max _{i=1}^{M} \max _{j \in \mathcal{N}_{i,k}^{(l)} } K_{i, j},
	\end{align*}
     where $k$ is that for $k$-NN used in defininig $N_i$ and
   \begin{align*}
    K_{i, j}=\max \left\{\frac{d^{(l)}_{i,j}}{d^{(l')}_{i,j}}, \frac{d^{(l')}_{i,j}}{d^{(l)}_{i,j}}\right\}.
	\end{align*}
 
    \item[(8)] \textbf{MRE}   (Mean reconstruction error) measures the difference between the input and output of an autoencoder, as usually defined. More generally, an MRE may also be defined to measure the difference between a pair of corresponding data in the multi-layer encoder and decoder.

\end{enumerate}

While the meanings of \textbf{MRE} and KL-Divergence are well known, those of the other metrics are explained as follows:

\begin{itemize}

\item {\bf \#Succ} is the primary metric measuring the success rate of unfolding a manifold. Without successful unfolding, the other metrics would not make sense. This metric is based on manual observation (see examples in A.8 at the end of this document).

\item {\bf RRE}, {\bf Trust} and {\bf Cont} all measure changes in neighboring relationships across-layers. We think that of these three, the Cont is more appropriate than the other two because it emphasizes on the neighboring relationship in the target latent embedding space. 

\item {\bf LGD} is the primary metric measuring the degree to which the LIS constraint is violated. However, it is unable to detect folding.

\item \textbf{$K$-min} and \textbf{$K$-max}  are the key metrics for the local bi-Lipschitz continuity,  {$K$-max} $\ge$ {$K$-min} $\ge 1$. The closer to 1 they are, the better the network homeomorphism preserves the isometry, the more stable the network is in training and the more robust is it against adversarial attacks. 

\item \textbf{$K$-max} can effectively identify the collapse in the latent space. This is because if the mapping maps two distinct input samples to an identical one (collapse), \textbf{$K$-max} will become huge (infinity) whereas {\bf L-KL} is not sensitive to such collapse.

\end{itemize}

Every set of experiments is run 10 times, each with a data set generated using a random seed in $\{0,1,\cdots,9\}$. Every final metric shown is the average of the 10 results. When a run is unsuccessful in unfolding the input manifold data, the resulting averaged statistics of metrics are not very meaningful so the numbers will be shown in \textcolor[rgb]{0.5,0.5,0.5}{gray color} in the following tables of evaluation metrics.



\subsection*{A.2 ML-Enc manifold learning}

Table~\ref{tab:S-CurveA3} compares the ML-Enc with other algorithms in the 8 evaluation metrics for the S-Curve. The ML-Enc performs the best for all but MPE. Note, however, that t-SNE and LLE failed to unfold the manifold hence their results should be considered as invalid even if t-SNE achieved the lowest MPE value due to its collapsing to a small cluster. Because LLE and t-SNE have zero success rate, their metrics do not make sense hence not compared in Table~\ref{tab:S-CurveA3}. The results show that ML-Enc performs significantly better than the others for all the metrics except for MPE. 

\begin{table*}[t]
    \centering
    \caption{Comparison in embedding quality for S-Curve (800 points)}
	\begin{tabular}{lllllllll}
		\toprule
		& \#Succ     & L-KL                & RRE              & Cont            & LGD               & $K$-min           & $K$-max         & MPE       \\ \midrule
		ML-Enc & \textbf{10} & \textbf{0.00737} & \textbf{0.000054} & \textbf{0.999994} & \textbf{0.000402} & \textbf{1.0017}   & \textbf{1.17}   & 0.0585    \\
		MLLE   & \textbf{10} & 0.00931          & 0.008893          & 0.989029          & 0.045243          & 7.2454            & 32.43           & 0.1722    \\
		HLLE   & \textbf{10} & 0.00971          & 0.008866          & 0.989082          & 0.045243          & 7.2341            & 31.69           & \textbf{0.0804}    \\
		LTSA   & \textbf{10} & 0.00971          & 0.008866          & 0.989082          & 0.045243          & 7.2341            & 31.69           & 0.0948    \\
		ISOMAP & \textbf{10} & 0.00931          & 0.001762          & 0.999233          & 0.023359          & 1.1148            & 13.36           & 0.0173    \\
		LLE    & 0 & \textcolor[rgb]{0.5,0.5,0.5}{0.05462}	& \textcolor[rgb]{0.5,0.5,0.5}{0.008055}		& \textcolor[rgb]{0.5,0.5,0.5}{0.991993}		& \textcolor[rgb]{0.5,0.5,0.5}{0.046917}	& \textcolor[rgb]{0.5,0.5,0.5}{8.4081}	& \textcolor[rgb]{0.5,0.5,0.5}{143.31}	& \textcolor[rgb]{0.5,0.5,0.5}{0.1445}   \\
		t-SNE  & 0        & \textcolor[rgb]{0.5,0.5,0.5}{0.01486}	& \textcolor[rgb]{0.5,0.5,0.5}{0.002071}		& \textcolor[rgb]{0.5,0.5,0.5}{0.997830}		& \textcolor[rgb]{0.5,0.5,0.5}{1.36791}	& \textcolor[rgb]{0.5,0.5,0.5}{14.6310}	& \textcolor[rgb]{0.5,0.5,0.5}{311.24}	& \textcolor[rgb]{0.5,0.5,0.5}{0.0412} 
		         \\ \bottomrule
	\end{tabular}
    \label{tab:S-CurveA3}
\end{table*}

\begin{figure}[!htb]
	\centering 
	\includegraphics[width = 2.5in]{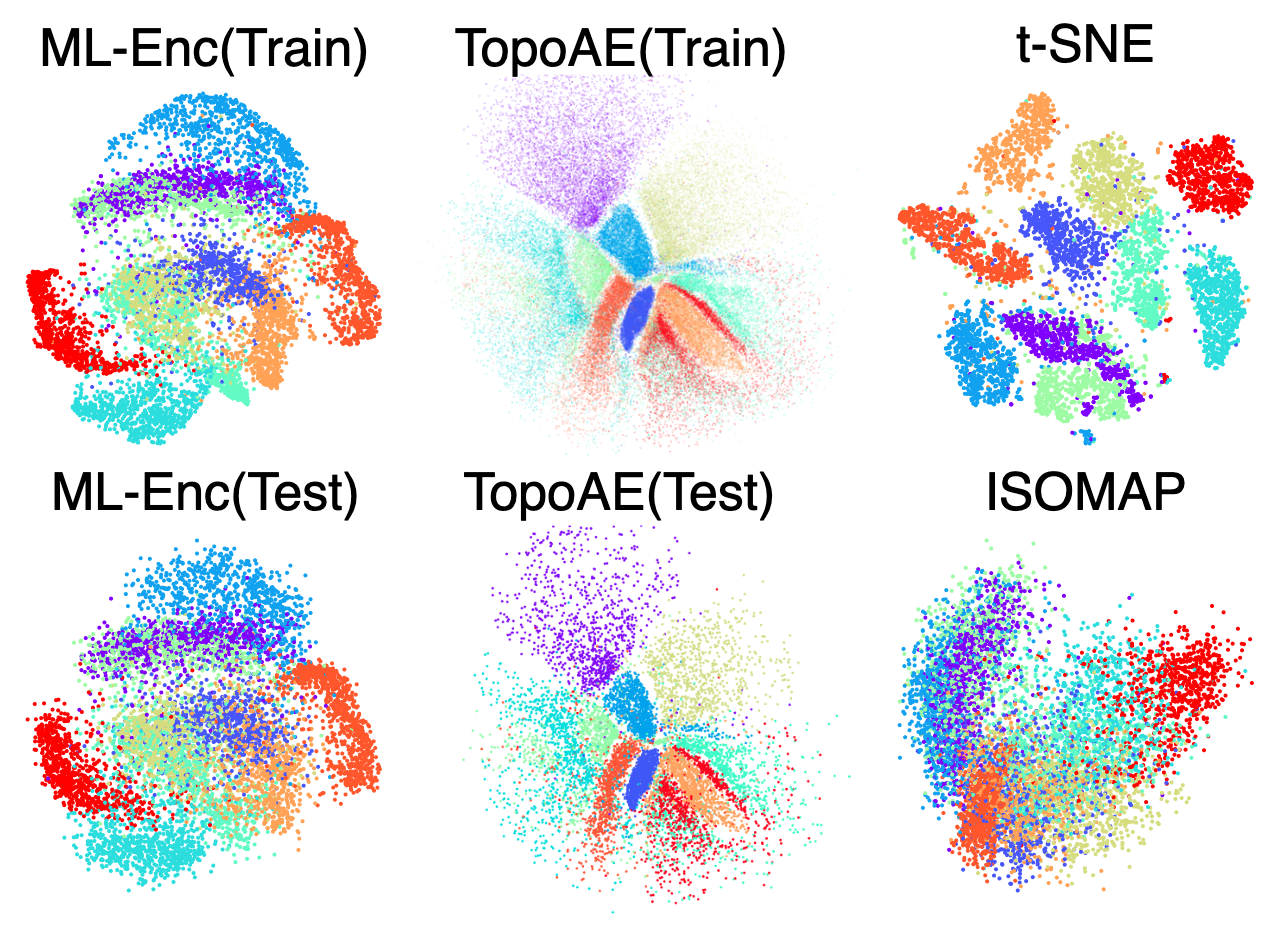}
	\caption{Visualization of MNIST results in 2D latent spaces.(The t-SNE and ISOMAP methods cannot be tested, so only the results of training are shown, same in Fig.~\ref{fig:Sphere10000} and Fig.~\ref{fig:Sphere5500})}
	\label{fig:mnist}
\end{figure}

\begin{figure}[!htb]
	\centering 
	\includegraphics[width = 2.5in]{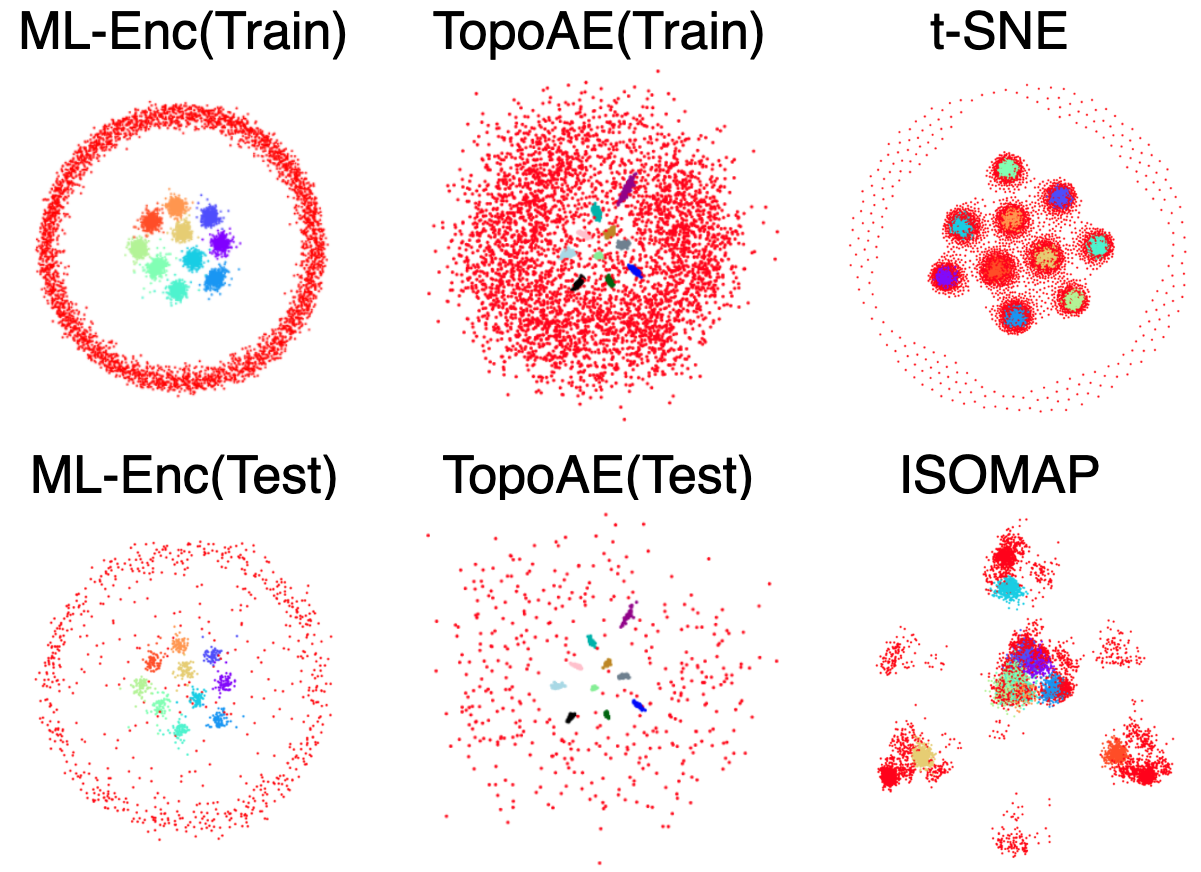}
	\caption{Visualization of Spheres10000 results in 2D latent spaces.}
	\label{fig:Sphere10000}
\end{figure}

\begin{figure}[!htb]
	\centering 
	\includegraphics[width = 2.5in]{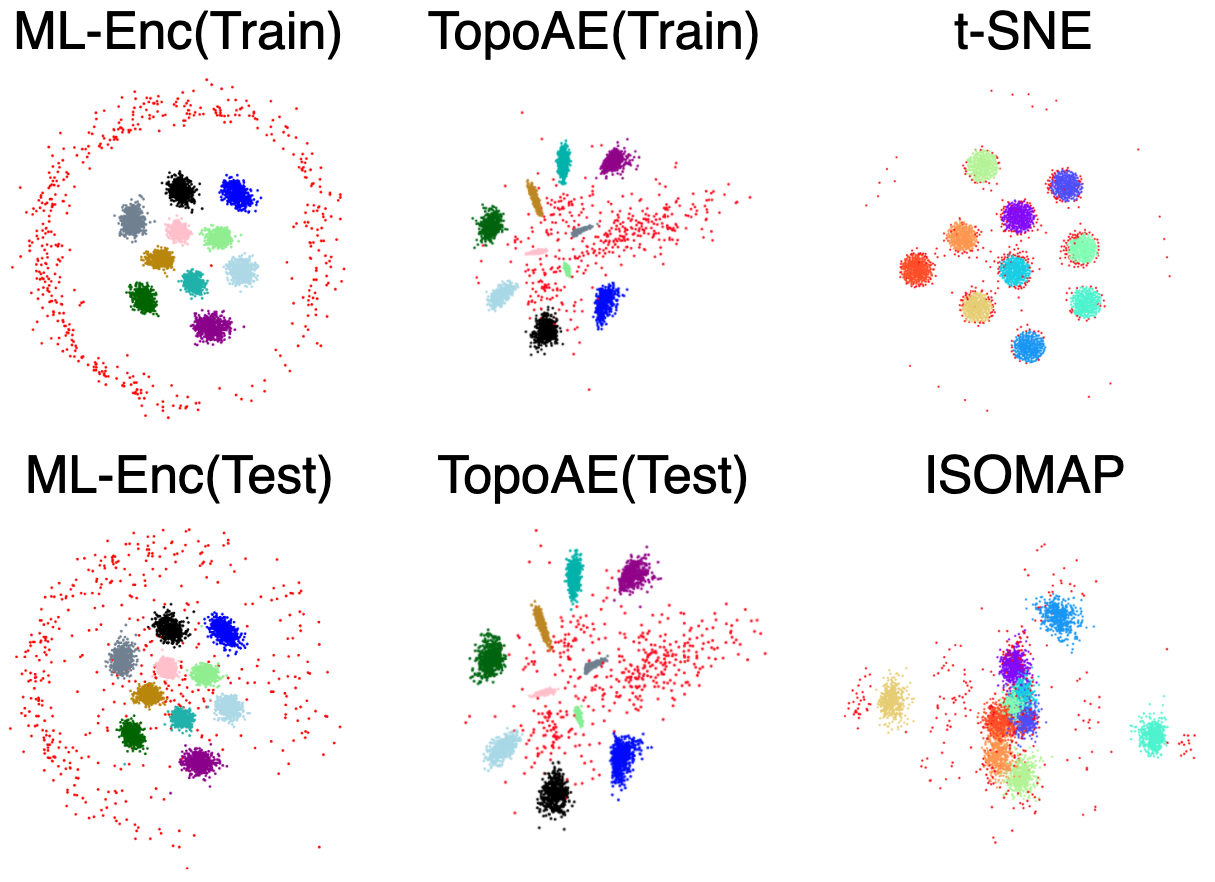}
	\caption{Visualization of Spheres5500 results in 2D latent spaces.}
	\label{fig:Sphere5500}
\end{figure}

Having evaluated toy datasets with perceivable structure, the following presents the results with the {\rm MNIST} dataset in 784-D and the {\bf Spheres} dataset in 101-D spaces, obtained by using ML-Enc and ML-AE  in comparison with others. For the {\bf MNIST} dataset of 10 digits, a subset of 8000 points are randomly chosen for training and another subset of 8000 points for testing without overlapping. The ML-Enc is used for NLDR by manifold learning. The results are shown in Fig.~\ref{fig:mnist}-\ref{fig:Sphere5500}.

\subsection*{A.3 Robustness to sparsity and noise}

Results for different simple sizes are presented in Table~\ref{tab:1} - \ref{tab:1.5}. Performance metrics under  different noise levels are presented in Table~\ref{tab:1.6} - \ref{tab:2}.Here, four metrics, \textbf{Trust}, \textbf{LGD}, $K$-\textbf{min} and $K$-\textbf{max},  are selected to compare ML-Enc with others.  

\begin{table}
	\begin{center}
	\caption{Cont with different sample sizes (given noise level = 0.0)}
	\begin{tabular}{l  l  l  l  l l}
	\toprule
	& 700 & 800 & 1000 & 1500 & 2000 \\
	\midrule
	ML-Enc & \textbf{0.9969} & \textbf{0.9984} & \textbf{0.9977} & \textbf{0.9999} & \textbf{0.9999} \\
	MLLE & 0.9760 & 0.9844 & 0.9891 & 0.9947 & 0.9960 \\
	HLLE & 0.9816 & 0.9851 & 0.9901 & 0.9947 & 0.9960 \\
	LTSA & 0.9816 & 0.9859 & 0.9900 & 0.9947 & 0.9960 \\
	ISOMAP & 0.9913 & 0.9950 & \textbf{0.9977} & 0.9995 & 0.9996\\
	\bottomrule
	\end{tabular}
	\label{tab:1}
	\end{center}
\end{table}

\begin{table}
\begin{center}
	\caption{LGD with different sample sizes (given noise level = 0.0)}
\begin{tabular}{l  l  l  l  l l}
\toprule
 & 700 & 800 & 1000 & 1500 & 2000 \\
 \midrule
ML-Enc & \textbf{0.0167} & \textbf{0.0040} & \textbf{0.0048} & \textbf{0.0016} & \textbf{0.0016} \\
MLLE & 0.0480 & 0.0453 & 0.0407 & 0.0335 & 0.0293 \\
HLLE & 0.0483 & 0.0454 & 0.0408 & 0.0335 & 0.0293 \\
LTSA & 0.0483 & 0.0454 & 0.0408 & 0.0335 & 0.0293 \\
ISOMAP & 0.0261 & 0.0238 & 0.0208 & 0.0161 & 0.0139 \\
\bottomrule
\end{tabular}
\end{center}
\end{table}

\begin{table}
\begin{center}
	\caption{$K$-min with different sample sizes (given noise level = 0.0)}
\begin{tabular}{l l l l l l}
\toprule
 & 700 & 800 & 1000 & 1500 & 2000 \\
\midrule
ML-Enc      & \textbf{1.0059}   & \textbf{1.0049} & \textbf{1.0167} & \textbf{1.0070} & \textbf{1.0063} \\
MLLE        & 6.4093            & 7.3723 & 8.5129 & 10.150 & 50.581 \\
HLLE        & 6.7227            & 7.4430 & 8.4837 & 10.149 & 48.563 \\
LTSA        & 6.7226            & 7.4428 & 8.4848 & 10.149 & 48.526 \\
ISOMAP      & 1.1069            & 1.1075 & 1.1026 & 1.0968 & 8.3342 \\
\bottomrule
\end{tabular}
\end{center}
\end{table}

\begin{table}
\begin{center}
	\caption{$K$-max with different sample sizes (given noise level = 0.0)}
\begin{tabular}{l l l l l l}
\toprule
 & 700 & 800 & 1000 & 1500 & 2000 \\
 \midrule
ML-Enc & \textbf{2.9917} & \textbf{1.7239}& \textbf{18.306} & \textbf{1.9931} & \textbf{2.1216} \\
MLLE & 360.03 & 240.62 & 442.48 & 43.901 & 50.581 \\
HLLE & 326.59 & 216.43 & 301.56 & 42.356 & 48.563 \\
LTSA & 314.81 & 215.84 & 296.37 & 42.358 & 48.526 \\
ISOMAP & 31.299 & 34.354 & 27.030 & 13.247 & 8.3342 \\
\bottomrule
\end{tabular}
\label{tab:1.5}
\end{center}
\end{table}


\begin{table}
\begin{center}
	\caption{Cont with different noise levels (given \#Samples = 800)}
\begin{tabular}{l  l  l  l  l  l  l}
\toprule
                  & 0.05     & 0.1      & 0.15     & 0.2      & 0.25     & 0.30     \\ 
\midrule
ML-Enc & \textbf{0.998} & \textbf{0.998} & \textbf{0.998} & \textbf{0.998} & \textbf{0.997} & \textbf{0.997} \\
MLLE & 0.986 & 0.986 & 0.984 & 0.984 & 0.983 & 0.981 \\
HLLE & 0.979 & 0.979 & 0.980 & 0.982 & 0.981 & 0.984 \\
LTSA & 0.979 & 0.979 & 0.980 & 0.982 & 0.981 & 0.984 \\
ISOMAP & 0.995 & 0.995 & 0.995 & 0.995 & 0.994 & 0.993 \\
\bottomrule
\end{tabular}
\label{tab:1.6}
\end{center}
\end{table}

\begin{table}
\begin{center}
	\caption{LGD with different noise levels (given \#Samples = 800)}
\begin{tabular}{l  l  l  l  l  l  l}
\toprule
                  & 0.05     & 0.1      & 0.15     & 0.2      & 0.25     & 0.30     \\ 
\midrule
ML-Enc & \textbf{0.004} & \textbf{0.004} & \textbf{0.004} & \textbf{0.005} & \textbf{0.006} & \textbf{0.006} \\
MLLE & 0.045 & 0.045 & 0.045 & 0.046 & 0.046 & 0.046 \\
HLLE & 0.045 & 0.045 & 0.046 & 0.046 & 0.047 & 0.047 \\
LTSA & 0.046 & 0.046 & 0.046 & 0.046 & 0.046 & 0.047 \\
ISOMAP & 0.023 & 0.023 & 0.023 & 0.023 & 0.023 & 0.024 \\
\bottomrule
\end{tabular}
\end{center}
\end{table}

\begin{table}
\begin{center}
	\caption{$K$-min with different noise levels (given \#Samples = 800)}
\begin{tabular}{l l l l l l l}
\toprule
                  & 0.05     & 0.1      & 0.15     & 0.2      & 0.25     & 0.30     \\
\midrule
ML-Enc & \textbf{1.005} & \textbf{1.005} & \textbf{1.006} & \textbf{1.006} & \textbf{1.009} & \textbf{1.011} \\
MLLE & 7.523 & 7.735 & 7.693 & 7.779 & 7.799 & 7.852 \\
HLLE & 6.977 & 7.018 & 6.987 & 7.233 & 8.016 & 8.328 \\
LTSA & 7.691 & 7.874 & 7.406 & 7.268 & 7.479 & 8.328 \\
ISOMAP & 1.114 & 1.114 & 1.125 & 1.117 & 1.104 & 1.097 \\
\bottomrule
\end{tabular}
\end{center}
\end{table}

\begin{table}
\begin{center}
\centering
\caption{$K$-max with different noise levels (given \#Samples = 800)}
\begin{tabular}{l l l l l l l}
\toprule
                  & 0.05     & 0.1      & 0.15     & 0.2      & 0.25     & 0.30     \\ 
\midrule
ML-Enc & \textbf{4.150} & \textbf{5.450} & \textbf{5.314} & \textbf{6.098} & \textbf{7.660} & \textbf{14.52} \\
MLLE & 304.8 & 181.1 & 256.5 & 416.2 & 272.5 & 299.8 \\
HLLE & 508.9 & 898.3 & 5997 & 8221 & 1162 & 714.1 \\
LTSA & 508.9 & 898.3 & 5997 & 8221 & 1162 & 714.1 \\
ISOMAP & 25.49 & 16.39 & 17.71 & 25.38 & 35.14 & 41.42 \\
\bottomrule
\end{tabular}
\label{tab:2}
\end{center}
\end{table}


\subsection*{A.4 Generalization to unseen data}

\begin{figure*}[t]
	\centering
	\includegraphics[width=0.95\linewidth]{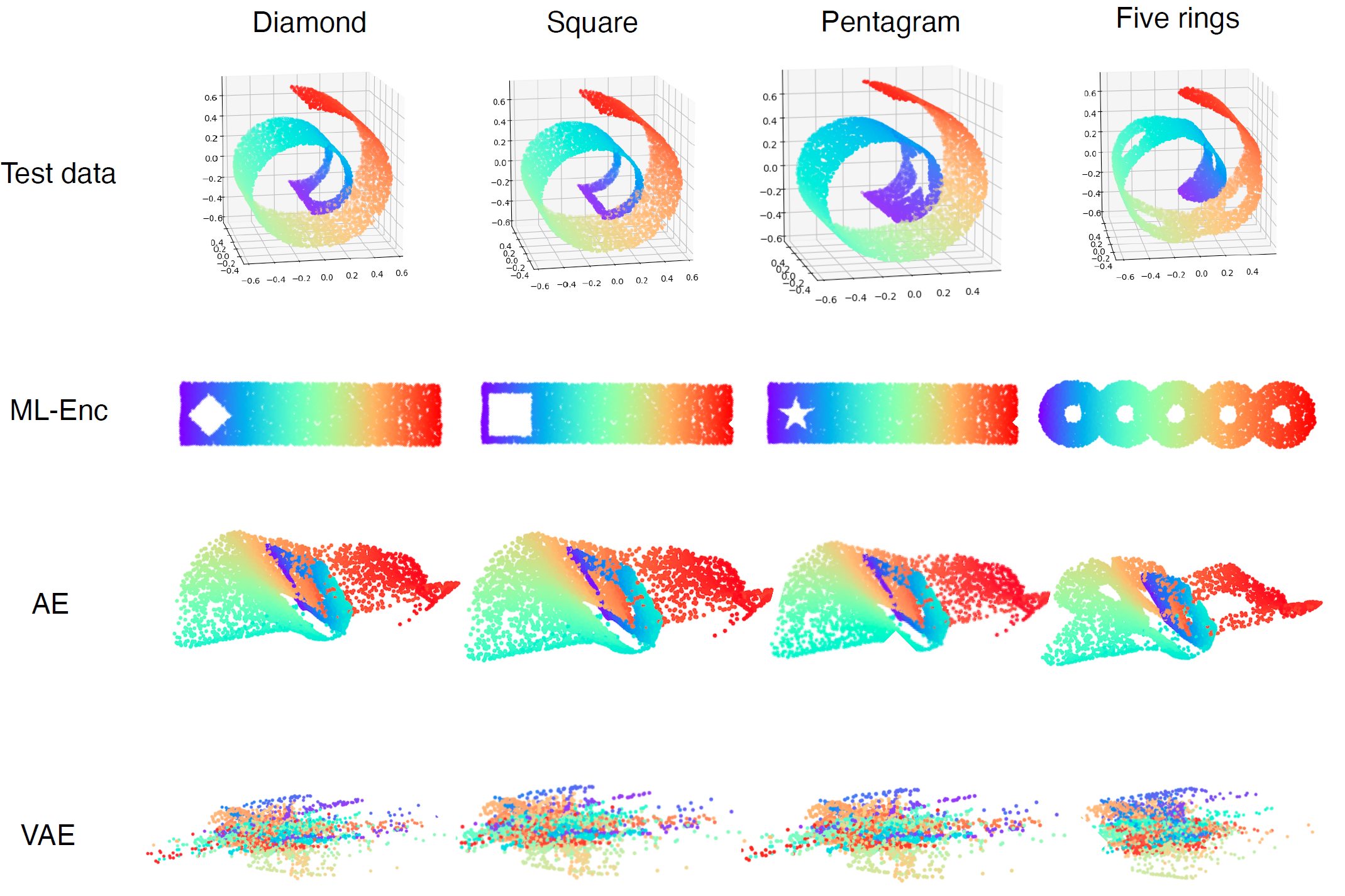}
	\caption{Ability of the learned ML-Enc network to generalize to unseen data for NLDR.}
	\label{fig:generalization}
\end{figure*}

Fig.~\ref{fig:generalization} demonstrates that a learned ML-Enc can generalize well to unseen data in unfolding a modified version of the same manifold to the corresponding version of embedding.  The ML-Enc model is trained  with a Swiss Roll (800 points) dataset. The test is done as follows: First, a set of 8000 points of the Swiss Roll manifold are generated; the data set is modified by removing, from the generated 8000 points of the manifold, the shape of a diamonds, square, pentagram or five-ring, respectively, creating 4 test sets. Each point  of a test set is transformed independently by the trained ML-Enc to obtain an embedding. We can see from each of the resulting embeddings that the unseen manifold data sets are well unfolded by the ML-Enc and the removed shapes are kept very well, illustrating that the learned ML-Enc has a good ability to generalize to unseen data. Since LLE-based, LTSA and ISOMAP algorithms do not possess such a generalization ability, the ML-Enc is compared with the encoder parts of the AE based algorithms. Unfortunately, AE and VAE failed altogether for the Swiss Roll data sets.

\subsection*{A.5 ML-AE for manifold data generation}

\begin{figure*}[t]
	\flushright
		\subfigure[training data]{ 
			\includegraphics[width=0.32\linewidth]{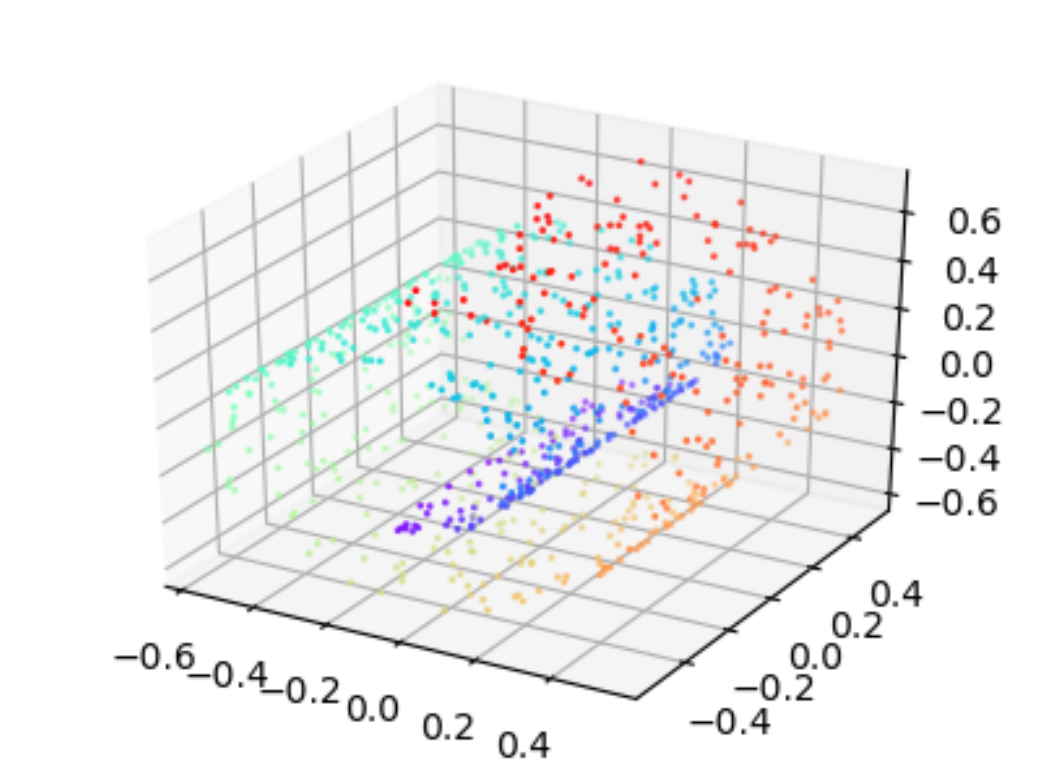}}
		\subfigure[embedding]{ 
			\includegraphics[width=0.32\linewidth]{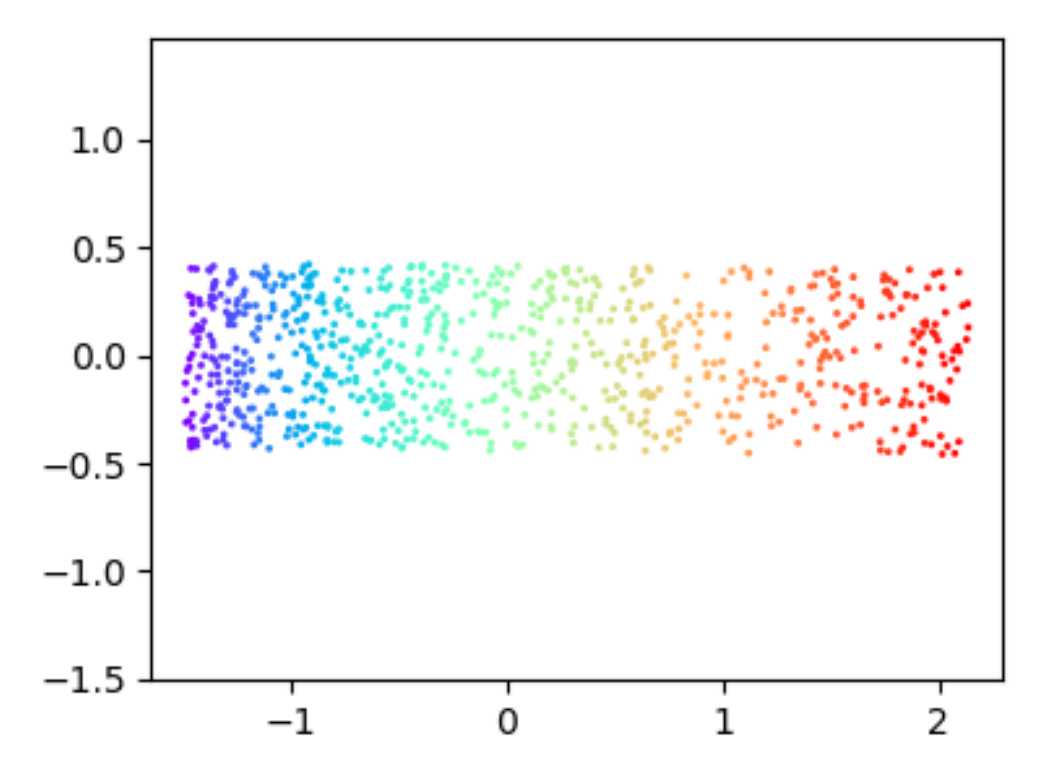}}
		\subfigure[reconstruction]{ 
			\includegraphics[width=0.32\linewidth]{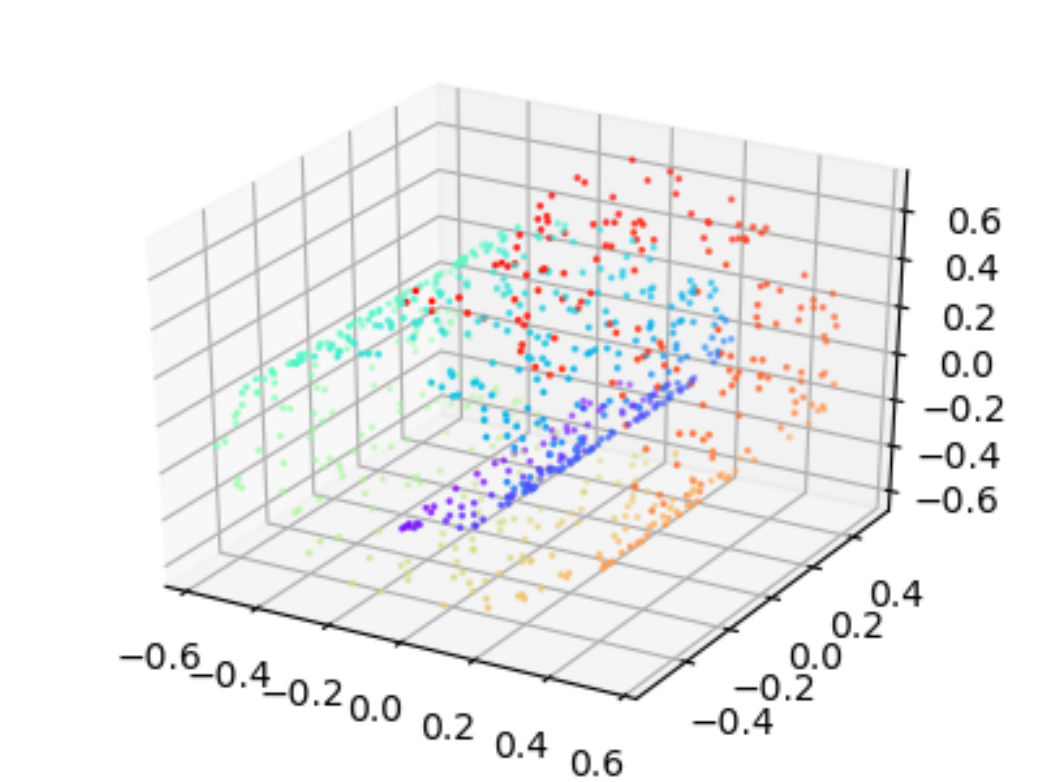}}
		\subfigure[random samples]{ 
			\includegraphics[width=0.32\linewidth]{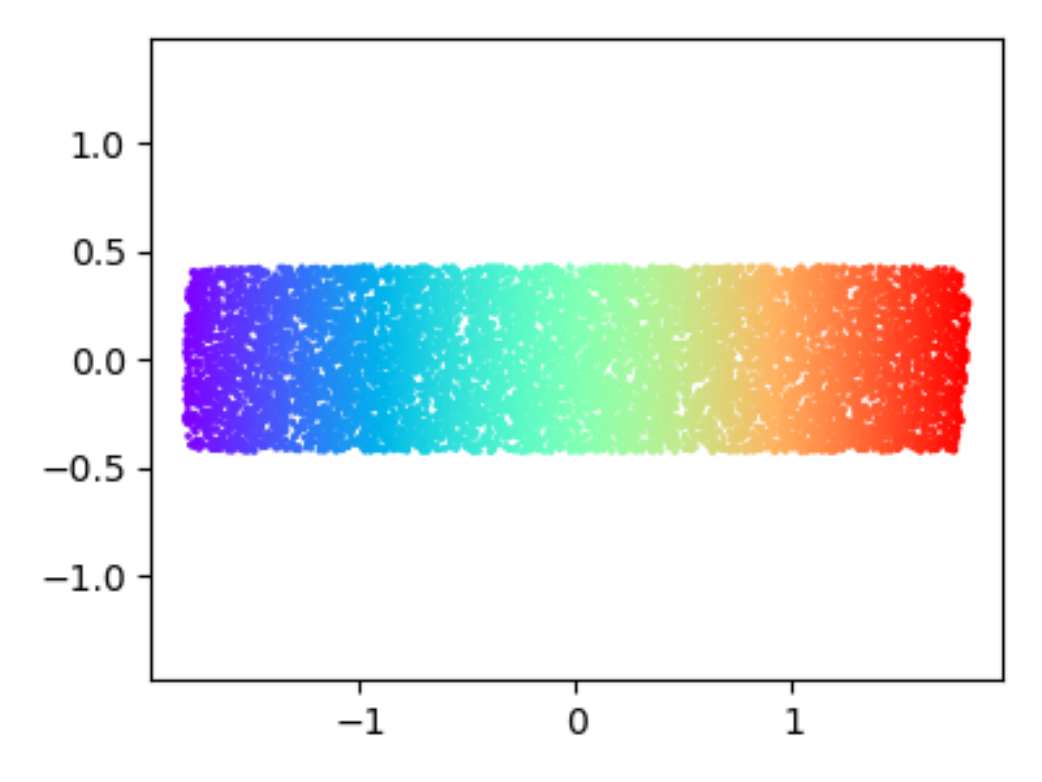}}
		\subfigure[generation]{ 
			\includegraphics[width=0.32\linewidth]{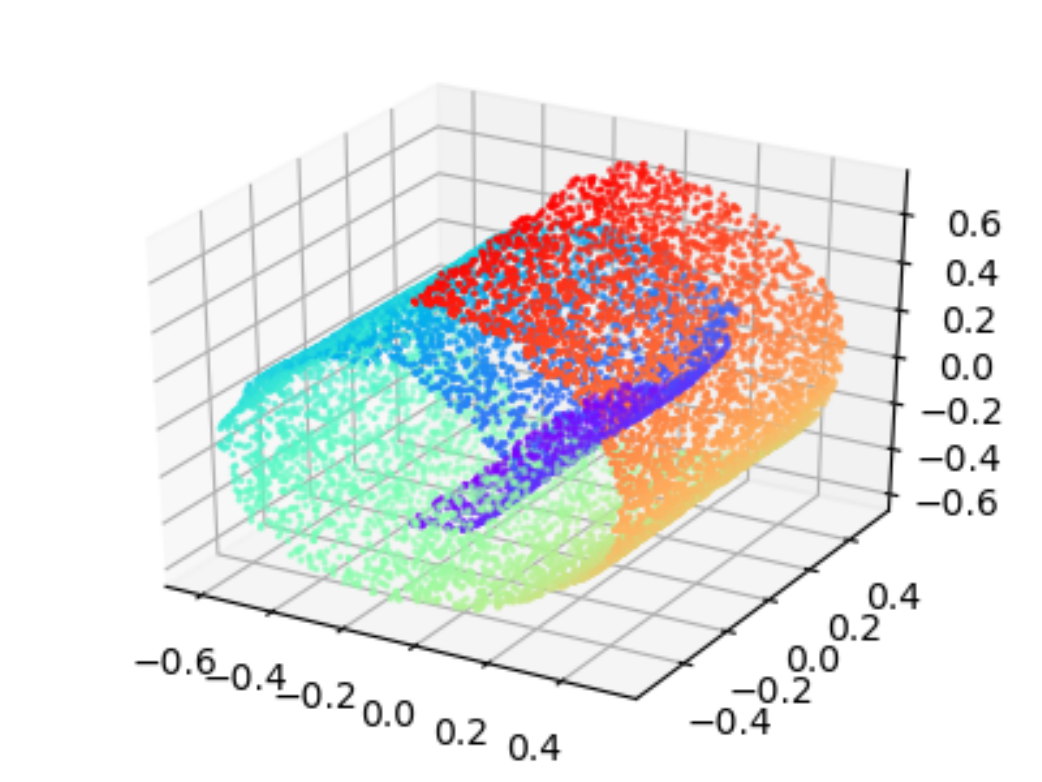}}
		\caption{ML-AE for manifold data generation.}
		\label{fig:manifolddatageneration}
\end{figure*}

Fig.~\ref{fig:manifolddatageneration}  shows manifold data reconstruction and generation using ML-AE, for which cases AE and VAE all failed in learning to unfold. In the learning phase, the ML-AE performs manifold learning for NLDR and then reconstruction, taking  (a) the training data in the ambient space as input, output embedding  (b) in the learned latent space, and then reconstruct back (c)  in the ambient data space. In the generation phase, the ML-Dec takes as random input samples (d) in the latent space, and maps the samples to the manifold (e) in the ambient data space.

\subsection*{A.6 More ablation about the ML-AE loss terms }

Three more sets of ablation experiments are provided, further demonstrating that the use of the LIS constraint in MLDL has accomplished its promises. The first set evaluate different \textbf{cross-layer weight schemes} for the ML-Enc, based on the 5-layer network architecture (3-100-100-100-3-2) presented in Section 4.1. The following 6 different cross-layer weight schemes (nonzero $\alpha$ weights in $\mathcal{L}_{lis}$) are evaluated:
	\begin{itemize}
	    \setlength{\itemindent}{2em}
		\item[M1:] $\alpha^{(0,5)}=1$ (between the input and latent layers only);
		\item[M2:] $\alpha^{(0,1)}=\frac{2}{30}$, $\alpha^{(1,2)}=\frac{4}{30}$, $\alpha^{(2,3)}=\frac{6}{30}$, $\alpha^{(3,4)}=\frac{8}{30}$, $\alpha^{(4,5)}=\frac{10}{30}$ (between each pair of adjacent layers, the weight increasing as the other layer goes deeper); 
		
		\item[M3:] $\alpha^{(1,5)}=\frac{2}{30}$, $\alpha^{(2,5)}=\frac{4}{30}$, $\alpha^{(3,5)}=\frac{6}{30}$, $\alpha^{(4,5)}=\frac{8}{30}$, $\alpha^{(0,5)}=\frac{10}{30}$ (between the latent layer and each of the other layers, the weight increasing as the other layer goes deeper); 
		
		\item[M4:] $\alpha^{(0,1)}=\frac{2}{30}$, $\alpha^{(0,2)}=\frac{4}{30}$, $\alpha^{(0,3)}=\frac{6}{30}$, $\alpha^{(0,4)}=\frac{8}{30}$, $\alpha^{(0,5)}=\frac{10}{30}$  (between the input layer and each of the other layers, the weight increasing as the other layer goes deeper).
		
		\item[M5:] $\alpha^{(0,1)}=\frac{1}{5}$, $\alpha^{(0,2)}=\frac{1}{5}$, $\alpha^{(0,3)}=\frac{1}{5}$, $\alpha^{(0,4)}=\frac{1}{5}$, $\alpha^{(0,5)}=\frac{1}{5}$  (between the input layer and each of the other layers, the weight being equal for all layers).
		
		\item[M6:] $\alpha^{(0,1)}=\frac{10}{30}$, $\alpha^{(0,2)}=\frac{8}{30}$, $\alpha^{(0,3)}=\frac{6}{30}$, $\alpha^{(0,4)}=\frac{4}{30}$, $\alpha^{(0,5)}=\frac{2}{30}$  (between the input layer and each of the other layers, the weight decreasing as the other layer goes deeper).
	\end{itemize}

Secondly, it is interesting to evaluate the ML-Enc in its ability to preserve the local geometry structure not only at the latent layer, but also at intermediate layers. To see this aspect, we provide the metrics calculated between layer $l=0$ and layer $l'\in \{1,3,5\}$. The ablation results are shown in Table~\ref{tab:cross-layer}. Schemes M1 and M4 have 100\% success rate for the 10 runs, M5 and M6 have some successes whereas M2 and M3 have zero. Of M1 and M4, the latter seems better overall. 

\begin{table*}
	\begin{center}
		\caption{Evaluation metrics with different cross-layer schemes for ML-Enc}
		\begin{tabular}{@{}llllllllll@{}}
			\toprule
			$l$-$l'$                  & Method & \#Succ      & L-KL            & RRE               & Cont           & LGD              & $K$-min & $K$-max & MPE             \\ \midrule
			\multirow{4}{*}{0-1} & M1     & \textbf{10}           & 0.0026          & 0.000792          & 0.9997          & 0.03081          & 2.29           & 3.29           & 0.0718               \\
			& M2     & 0           & \textcolor[rgb]{0.7,0.7,0.7}{0.0062}	& \textcolor[rgb]{0.7,0.7,0.7}{0.001156}	& \textcolor[rgb]{0.7,0.7,0.7}{0.9993}	& \textcolor[rgb]{0.7,0.7,0.7}{0.02981}	& \textcolor[rgb]{0.7,0.7,0.7}{2.06}	& \textcolor[rgb]{0.7,0.7,0.7}{3.45}  & \textcolor[rgb]{0.7,0.7,0.7}{0.0068}             \\
			& M3     & 0          & \textcolor[rgb]{0.7,0.7,0.7}{0.0038}	& \textcolor[rgb]{0.7,0.7,0.7}{0.001781}	& \textcolor[rgb]{0.7,0.7,0.7}{0.9990}	& \textcolor[rgb]{0.7,0.7,0.7}{0.02761}	& \textcolor[rgb]{0.7,0.7,0.7}{1.86}	& \textcolor[rgb]{0.7,0.7,0.7}{3.39}   & \textcolor[rgb]{0.7,0.7,0.7}{0.0093}            \\        
			& M4     & \textbf{10}           & \textbf{0.0011}          & \textbf{0.000555}          & \textbf{0.9998}          & \textbf{0.01494}          & \textbf{1.31}           & \textbf{1.72}           & \textbf{0.0135}               \\
			& M5     & 5           & \textcolor[rgb]{0.7,0.7,0.7}{0.0006}	& \textcolor[rgb]{0.7,0.7,0.7}{0.000220}	& \textcolor[rgb]{0.7,0.7,0.7}{0.9998}	& \textcolor[rgb]{0.7,0.7,0.7}{0.00199}	& \textcolor[rgb]{0.7,0.7,0.7}{1.00}	& \textcolor[rgb]{0.7,0.7,0.7}{1.18}  & \textcolor[rgb]{0.7,0.7,0.7}{0.0576}             \\
			& M6    & 2           & \textcolor[rgb]{0.7,0.7,0.7}{0.0003}	& \textcolor[rgb]{0.7,0.7,0.7}{0.000084}	& \textcolor[rgb]{0.7,0.7,0.7}{0.9999}	& \textcolor[rgb]{0.7,0.7,0.7}{0.00081}	& \textcolor[rgb]{0.7,0.7,0.7}{1.00}	& \textcolor[rgb]{0.7,0.7,0.7}{1.08}   & \textcolor[rgb]{0.7,0.7,0.7}{0.0566}            \\   
			\midrule
			\multirow{4}{*}{0-3} & M1     & \textbf{10}           & 0.0235          & 0.001534          & 0.9980          & 0.03541          & 2.28           & 5.98           & 0.0718               \\
			& M2    & 0           & \textcolor[rgb]{0.7,0.7,0.7}{0.0667}	& \textcolor[rgb]{0.7,0.7,0.7}{0.047127}	& \textcolor[rgb]{0.7,0.7,0.7}{0.9620}	& \textcolor[rgb]{0.7,0.7,0.7}{0.03642}	& \textcolor[rgb]{0.7,0.7,0.7}{1.98}	& \textcolor[rgb]{0.7,0.7,0.7}{43.83}  & \textcolor[rgb]{0.7,0.7,0.7}{0.0068}             \\
			& M3    & 0           & \textcolor[rgb]{0.7,0.7,0.7}{0.0292}	& \textcolor[rgb]{0.7,0.7,0.7}{0.053970}	& \textcolor[rgb]{0.7,0.7,0.7}{0.9802}	& \textcolor[rgb]{0.7,0.7,0.7}{0.02235}	& \textcolor[rgb]{0.7,0.7,0.7}{1.09}	& \textcolor[rgb]{0.7,0.7,0.7}{49.98}   & \textcolor[rgb]{0.7,0.7,0.7}{0.0093}            \\   
			& M4     & \textbf{10}           & \textbf{0.0135}          & \textbf{0.000516}          & \textbf{0.9986} & \textbf{0.00443}          & \textbf{1.00}  & \textbf{1.57}           & \textbf{0.0135}               \\ 
			& M5     & 5           & \textcolor[rgb]{0.7,0.7,0.7}{0.0024}	& \textcolor[rgb]{0.7,0.7,0.7}{0.000182}	& \textcolor[rgb]{0.7,0.7,0.7}{0.9998}	& \textcolor[rgb]{0.7,0.7,0.7}{0.00110}	& \textcolor[rgb]{0.7,0.7,0.7}{1.00}	& \textcolor[rgb]{0.7,0.7,0.7}{1.14}   & \textcolor[rgb]{0.7,0.7,0.7}{0.0576}            \\
			& M6     & 2          & \textcolor[rgb]{0.7,0.7,0.7}{0.0031}	& \textcolor[rgb]{0.7,0.7,0.7}{0.000216}	& \textcolor[rgb]{0.7,0.7,0.7}{0.9998}	& \textcolor[rgb]{0.7,0.7,0.7}{0.00100}	& \textcolor[rgb]{0.7,0.7,0.7}{1.00}	& \textcolor[rgb]{0.7,0.7,0.7}{1.14}   & \textcolor[rgb]{0.7,0.7,0.7}{0.0566}            \\    \midrule
			\multirow{4}{*}{0-5} & M1     & \textbf{10} & \textbf{0.0184} & \textbf{0.000414} & \textbf{0.9998} & \textbf{0.00385} & \textbf{1.00}  & 2.14           & 0.0718          \\
			& M2    & 0           & \textcolor[rgb]{0.7,0.7,0.7}{0.0655}	& \textcolor[rgb]{0.7,0.7,0.7}{0.102837}	& \textcolor[rgb]{0.7,0.7,0.7}{0.9499}	& \textcolor[rgb]{0.7,0.7,0.7}{0.03631}	& \textcolor[rgb]{0.7,0.7,0.7}{1.71}	& \textcolor[rgb]{0.7,0.7,0.7}{1152.42}	& \textcolor[rgb]{0.7,0.7,0.7}{0.0068}               \\
			& M3     & 0           & \textcolor[rgb]{0.7,0.7,0.7}{0.0488}	& \textcolor[rgb]{0.7,0.7,0.7}{0.105081}	& \textcolor[rgb]{0.7,0.7,0.7}{0.9661}	& \textcolor[rgb]{0.7,0.7,0.7}{0.02166}	& \textcolor[rgb]{0.7,0.7,0.7}{1.09}	& \textcolor[rgb]{0.7,0.7,0.7}{994.12}	& \textcolor[rgb]{0.7,0.7,0.7}{0.0093}               \\        
			& M4     & \textbf{10} & \textbf{0.0184} & 0.000440          & 0.9984 & 0.00400          & \textbf{1.00}  & \textbf{1.73}  & \textbf{0.0135}          \\ 
			& M5     & 5          & \textcolor[rgb]{0.7,0.7,0.7}{0.0327}	& \textcolor[rgb]{0.7,0.7,0.7}{0.022078}	& \textcolor[rgb]{0.7,0.7,0.7}{0.9969}	& \textcolor[rgb]{0.7,0.7,0.7}{0.00630}	& \textcolor[rgb]{0.7,0.7,0.7}{1.01}	& \textcolor[rgb]{0.7,0.7,0.7}{3.68}	& \textcolor[rgb]{0.7,0.7,0.7}{0.0576}               \\
			& M6     & 2           & \textcolor[rgb]{0.7,0.7,0.7}{0.0454}	& \textcolor[rgb]{0.7,0.7,0.7}{0.030612}	& \textcolor[rgb]{0.7,0.7,0.7}{0.9942}	& \textcolor[rgb]{0.7,0.7,0.7}{0.01321}	& \textcolor[rgb]{0.7,0.7,0.7}{1.05}	& \textcolor[rgb]{0.7,0.7,0.7}{8.13}	& \textcolor[rgb]{0.7,0.7,0.7}{0.0566}               \\   \bottomrule
		\end{tabular}
		\label{tab:cross-layer}
	\end{center}
\end{table*}
	
Thirdly, we provide ablation experiments with different \textbf{corresponding-layer weight schemes for the ML-AE}, based on the ML-AE  architecture (3-100-100-100-3-2-3-100-100-100-100-3) presented in Section 4.3 where the LIS is imposed between layers 0 and 5 with $\alpha^{(0,5)}=1$ and all $\gamma^{(l,l')}=0.2$. A corresponding-layer scheme is determined by nonzero $\alpha$ weights in $\mathcal{L}_{lis}$ between the corresponding layers in the ML-Enc and ML-Dec. The following 4 weight schemes are evaluated:
	\begin{itemize}
	\setlength{\itemindent}{2em}
		\item[M7:] $\alpha^{(0,5)}=1$ (No LIS constraints between corresponding-layers, as baseline); 
		
		\item[M8:]  $\alpha^{(0,5)}=1$, $\alpha^{(0,0')}=\frac{2}{30}$, $\alpha^{(1,1')}=\frac{4}{30}$, $\alpha^{(2,2')}=\frac{6}{30}$, $\alpha^{(3,3')}=\frac{8}{30}$, $\alpha^{(4,4')}=\frac{10}{30}$ (the corresponding-layer weight for the LIS constraint increases as the layer number becomes bigger);
		
		\item[M9:]  $\alpha^{(0,5)}=1$, $\alpha^{(0,0')}=\frac{1}{5}$, $\alpha^{(1,1')}=\frac{1}{5}$, $\alpha^{(2,2')}=\frac{1}{5}$, $\alpha^{(3,3')}=\frac{1}{5}$, $\alpha^{(4,4')}=\frac{1}{5}$ (all the corresponding-layer weights for the LIS constraint are equal); 
		
		\item[M10:] $\alpha^{(0,5)}=1$, $\alpha^{(0,0')}=\frac{10}{30}$, $\alpha^{(1,1')}=\frac{8}{30}$, $\alpha^{(2,2')}=\frac{6}{30}$, $\alpha^{(3,3')}=\frac{4}{30}$, $\alpha^{(4,4')}=\frac{2}{30}$ (the corresponding-layer weight for the LIS constraint decreases as the layer number becomes bigger).
		
	\end{itemize}
The performance metrics are shown in Table~\ref{tab:corresponding-layerAE}, where the metric numbers are calculated between layers $l=0$ and $l'=0'$. The results demonstrate that M10 outperforms the other three schemes in all metrics except for one $K$-max. When compared with incremental and equal weight scheme (M8 and M9), decreasing the corresponding-layer weight for the LIS constraint as the layer number becomes bigger can result in the greater performance gain, since the closer the data is to the input layer in the Encoder, the more authentic and reliable it is.

	\begin{table*}
	\begin{center}
		\caption{ Evaluation metrics with different corresponding-layer schemes for ML-AE}
		\resizebox{\textwidth}{!}{
			\begin{tabular}{@{}lllllllllll@{}}
				\toprule
				 & & \#Succ      & L-KL             & RRE              & Cont           & LGD              &$K$-min & $K$-max & MPE              & MRE              \\ \midrule
				 & M7     & \textbf{10} & 0.00165          & 0.00070          & \textbf{0.9998} & 0.00514          & \textbf{1.017} & 2.540          & 0.04309          & 0.01846          \\
				& M8     & \textbf{10} & 0.00142          & 0.00065          & \textbf{0.9998} & 0.00480          & \textbf{1.017} & \textbf{2.324} & 0.04288          & 0.01810          \\
				& M9    & \textbf{10} & 0.00168          & 0.00070          & \textbf{0.9998} & 0.00520          & 1.018          & 2.543          & 0.04279          & 0.01788          \\
				& M10    & \textbf{10} & \textbf{0.00121} & \textbf{0.00063} & \textbf{0.9998} & \textbf{0.00459} & \textbf{1.017} & 2.377          & \textbf{0.04269} & \textbf{0.01762} \\ \bottomrule
		\end{tabular}}
		\label{tab:corresponding-layerAE}
	\end{center}
	\end{table*}

\end{document}